\documentclass{article}

\usepackage{PRIMEarxiv}

\usepackage[utf8]{inputenc} 
\usepackage[T1]{fontenc}    
\usepackage{hyperref}       
\usepackage{url}            
\usepackage{booktabs}       
\usepackage{amsfonts}       
\usepackage{nicefrac}       
\usepackage{microtype}      
\usepackage{lipsum}
\usepackage{fancyhdr}       
\usepackage{graphicx}       
\graphicspath{{media/}}     

\graphicspath{ {./figures/} }
\usepackage{float}
\usepackage{verbatim} 
\restylefloat{figure}
\floatstyle{plaintop} 
\restylefloat{table}
\usepackage{amsmath}
\usepackage{amssymb}
\usepackage{booktabs}
\usepackage{makecell}
\usepackage{multirow}
\usepackage{graphicx}
\usepackage{natbib}

\title{MATANet: A Multi-context Attention and Taxonomy-Aware Network for Fine-Grained Underwater Recognition of Marine Species}

\author{
  \textbf{Donghwan Lee} \quad \textbf{Byeongjin Kim} \quad \textbf{Geunhee Kim} \\
  \textbf{Nahyeon Maeng} \quad \textbf{Hyukjin Kwon} \quad \textbf{Wooju Kim}\thanks{Corresponding author.} \\
  Department of Industrial Engineering \\
  Yonsei University \\
  Seoul 03722, Republic of Korea \\
  \texttt{dhlee.ie@yonsei.ac.kr, jin\_kbj@yonsei.ac.kr} \\
  \texttt{rmsgml8689@gmail.com, mnh7075@yonsei.ac.kr} \\
  \texttt{sandiegokwondan@yonsei.ac.kr, wkim@yonsei.ac.kr}
}

\begin{document}
\maketitle

\begin{abstract}
Accurate fine-grained recognition of marine organisms is important for scalable biodiversity monitoring and ecological assessment using underwater imagery. However, existing methods mainly focus on target appearance and make limited use of surrounding environmental cues and biological taxonomy. We propose the Multi-Context Attention and Taxonomy-Aware Network (MATANet) for region-of-interest (ROI)-guided marine organism recognition. MATANet contains two complementary components. The Multi-Context Environmental Attention Module uses the ROI representation as a query to aggregate spatial patch features from ROI-centered contextual views at multiple scales, enabling target-conditioned modeling of the surrounding environment. Level-wise auxiliary classifiers further incorporate higher taxonomic ranks during training, encouraging hierarchically consistent representations without changing the finest-label prediction space or inference procedure. On the official FathomNet 2025 Private test split, MATANet achieves a hierarchical distance of 1.570 with the base backbone and 1.423 with the large backbone, substantially outperforming the strongest benchmark value of 2.603. On FishCLEF2015, MATANet achieves an accuracy of 0.793 and a hierarchical distance of 1.120, outperforming the strongest benchmark values of 0.766 and 1.327, respectively. Ablation studies show that surrounding scene information provides complementary evidence beyond repeated multi-scale observations of the target and that target-conditioned aggregation outperforms direct multi-view concatenation. Additional experiments on FAIR1M v2.0 examine the applicability of the proposed design beyond underwater imagery. In post-detection evaluation, MATANet improves fine-grained classification accuracy on matched detector-generated ROIs from 0.828 to 0.959 without additional fine-tuning, supporting its engineering applicability to automated marine monitoring.
\end{abstract}


\section{Introduction}
\label{introduction}

Accurate recognition of marine organisms in underwater digital imagery is essential for biodiversity management, environmental assessment, and evidence-based conservation planning \citep{RN1}. Marine ecosystem monitoring increasingly relies on large-scale underwater imagery acquired using diverse imaging platforms, including remotely operated vehicles, time-lapse stations, and towed camera systems \citep{RN401}. These imaging systems enable repeated and wide-area observations of marine environments, but they also produce rapidly growing volumes of visual data that are difficult to analyze manually \citep{RN12}. Manual taxonomic identification remains time-consuming and costly \citep{RN2}, and its scalability is further constrained by the shortage of trained taxonomic experts \citep{RN3}. These challenges underscore the need for automated systems that can support accurate and scalable taxonomic analysis in practical marine monitoring pipelines.

In real-world underwater imagery, marine organisms often appear as localized targets within complex underwater scenes \citep{RN12}. Therefore, in practical marine monitoring pipelines, automated recognition can be decomposed into target localization and subsequent fine-grained classification. Although deep learning-based object detection methods can provide candidate regions containing marine organisms \citep{RN103,RN104,RN105}, their detection-oriented classification components may not sufficiently capture the subtle visual differences required for fine-grained discrimination \citep{RN106,RN107}. Accordingly, a dedicated fine-grained visual classification (FGVC) stage can complement object detection by enabling more precise taxonomic recognition from localized target regions. Specifically, marine organism FGVC can be formulated as a region-of-interest (ROI)-guided classification problem, where the target organism is specified by a detected or annotated region and the objective is to predict its finest available taxonomic label. 

\begin{figure}[t]
  \centering \includegraphics[width=0.7\linewidth]{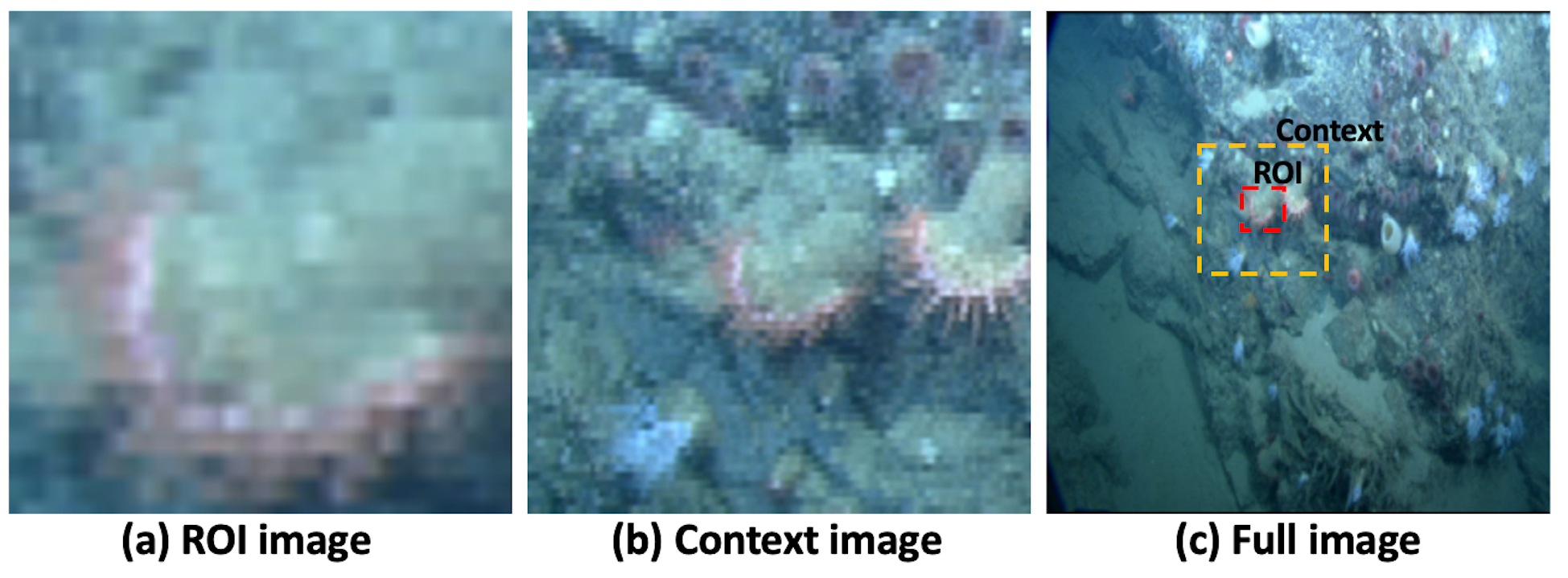}
  \caption{Visual contexts at different spatial scales for marine organism recognition. The ROI image in (a) shows a sea anemone from the family Hormathiidae. The nearby organisms and habitat features visible in the local-context image (b) and the full image (c) may provide additional cues for taxonomic identification.}
   \label{fig:challenge}
\end{figure}

Early deep learning studies on marine organism classification primarily focused on fish recognition and relied on convolutional neural network (CNN)-based representations, often using independently curated datasets or relatively coarse taxonomic categories \citep{RN5}. More recent research has expanded toward fine-grained marine organism recognition through the construction of larger-scale datasets \citep{RN11,RN12} and the development of transformer-based models designed to capture subtle visual differences \citep{RN14,RN15,RN17}. Together, these advances have improved visual representation learning for fine-grained marine organism recognition.

Despite these advances, two aspects remain insufficiently explored in fine-grained marine organism recognition. First, many existing methods primarily rely on target-centered visual representations to characterize individual organisms \citep{RN8,RN22,RN15,RN17}, as illustrated by the Hormathiidae ROI in Figure~\ref{fig:challenge}(a). However, focusing primarily on the target region may overlook relevant information from the surrounding environment. This issue is particularly important in underwater scenes, where constrained visibility, scale variation, and background clutter complicate fine-grained recognition \citep{RN14}. For example, as shown in Figure~\ref{fig:challenge}(b) and (c), nearby organisms, substrates, and habitat structures surrounding the Hormathiidae ROI may provide complementary evidence for taxonomic identification. Such contextual information, however, is not explicitly or selectively exploited by many conventional target-centered approaches. Second, many fine-grained marine organism recognition methods are trained using only the finest-level class labels, leaving relationships across taxonomic ranks largely unused as supervisory signals during representation learning. Biological taxonomy provides a structured hierarchy, and closely related taxa often share morphological characteristics \citep{RN306}. Conventional flat-label supervision treats fine-grained categories as independent labels and does not explicitly encode their hierarchical relationships or degrees of taxonomic relatedness. Moreover, visual similarity does not necessarily correspond to taxonomic proximity: taxonomically distant organisms may exhibit similar morphology, while shared habitats and imaging conditions may further reduce the reliability of appearance-based cues. Consequently, taxonomy-aware supervision may encourage the model to learn representations that better reflect the underlying biological structure.

In this study, we propose MATANet, a Multi-Context Attention and Taxonomy-Aware Network for ROI-guided fine-grained marine organism recognition. The design is motivated by the complementary roles of target appearance, surrounding habitat cues, and biological taxonomy in organism identification. Given a specified target ROI, the model constructs ROI-centered contextual views at progressively broader spatial scopes and encodes all views using a shared visual backbone. The proposed Multi-Context Environmental Attention Module (MCEAM) then uses the ROI representation as a semantic query to aggregate spatially resolved patch features from each contextual view. This formulation allows contextual information to be selected according to its interaction with the specified target representation while retaining the original ROI representation for fine-grained discrimination. In addition, MATANet incorporates biological taxonomy as training-only auxiliary supervision. Separate level-wise classifiers predict the higher taxonomic labels and guide the fused representation toward a taxonomically consistent structure, while the main classifier continues to operate in the original finest-level label space.

Experiments on FathomNet 2025~\citep{RN19} and FishCLEF2015~\citep{RN21} demonstrate that MATANet consistently outperforms representative benchmark models for fine-grained marine organism recognition. On the official FathomNet 2025 Private test split, MATANet achieves an HD of 1.570 with the ViT-B/14 backbone, compared with 2.603 for the strongest benchmark model. On FishCLEF2015, MATANet achieves an accuracy of 0.793 and an HD of 1.120, outperforming the strongest benchmark results of 0.766 and 1.327, respectively. The ablation studies examine the contributions of multi-scale target observations, surrounding scene content, contextual aggregation, and taxonomy-aware supervision. The results indicate that both repeated target observations and surrounding scene information contribute to recognition, with scene context providing particularly important complementary evidence. Under the same contextual views, backbone, and supervision configuration, MCEAM achieves better mean performance than direct concatenation across all reported metrics. Hierarchy-aware auxiliary objectives, including level-wise taxonomic supervision, improve hierarchical consistency while preserving flat classification performance, whereas directly replacing the main fine-grained CE loss with hierarchy-aware losses provides limited benefit. Post-detection experiments further show that the ViT-L/14 variant improves classification accuracy from 0.828 to 0.959 on detector-generated ROIs without additional fine-tuning. We additionally evaluate MATANet on FAIR1M v2.0~\citep{RN20} to examine the proposed multi-context and hierarchy-aware design in a distinct non-marine visual domain. The main contributions of this study are summarized as follows:

\begin{itemize}

\item We formulate contextual reasoning in ROI-guided fine-grained recognition as target-conditioned aggregation and introduce the Multi-Context Environmental Attention Module (MCEAM), which uses the specified ROI representation as a query to aggregate spatially resolved patch features from multiple ROI-centered contextual views.

\item We incorporate biological taxonomy through level-wise auxiliary classifiers used only during training, encouraging taxonomically consistent representations while preserving the original finest-level prediction space and inference procedure.

\item We provide extensive experimental validation on FathomNet 2025 and FishCLEF2015, together with post-detection evaluation using detector-generated ROIs and additional experiments on FAIR1M v2.0 beyond marine imagery.

\end{itemize}

\section{Related Work}
\subsection{Fine-Grained Marine Organism Recognition}

Early studies on marine organism classification often focused on fish recognition and mainly relied on convolutional neural network (CNN) architectures~\citep{RN22}. For example, \citet{RN8} combined CNN-based feature extraction with a linear SVM classifier to recognize 23 fish categories, demonstrating the potential of deep visual representations for underwater organism classification. More recent studies have shifted toward fine-grained visual classification (FGVC) to address subtle taxonomic differences among marine organisms. In this line of work, \citet{RN15} proposed the Multi-Scale Dual-Branch Network (MSDBN), which combines Vision Transformer (ViT) and CNN components to improve fine-grained feature representation. Similarly, \citet{RN17} introduced SwinFishNet, a Swin Transformer-based model trained with transfer learning, and reported competitive performance across multiple fish datasets. Recent studies have also explored hierarchical and multi-scale feature modeling for broader marine organism recognition. For instance, \citet{RN102} proposed a hierarchical multi-scale attention-based CNN built on an EfficientNetV2 backbone, achieving lightweight and accurate marine organism classification under varying underwater conditions. Despite these advances, existing marine organism recognition methods, including recent FGVC models, primarily rely on target-centered visual representations~\citep{RN8,RN15,RN17,RN102}. While these methods effectively capture fine-grained morphological traits, the role of surrounding environmental context remains relatively underexplored in marine FGVC. Accordingly, we investigate whether target-conditioned contextual information can complement target-centered representations in ROI-guided marine organism recognition.

\subsection{Multi-Stream Networks and Context Modeling}
Multi-stream network architectures process multiple inputs through parallel branches and fuse their representations for downstream prediction. Such architectures have been widely studied for integrating complementary visual information. In this line of research, \citet{RN23} separated the ROI and the surrounding background into independent networks and combined their representations using a multilayer perceptron (MLP), aiming to balance foreground discrimination and contextual cues. Similarly, \citet{RN24} improved disease classification performance by introducing a cross-attention-based fusion mechanism between representations extracted by two identical image encoders. Multi-stream designs have also been actively explored in medical imaging, where multi-scale information plays an important role. For example, \citet{RN25} proposed Deep-Hipo for whole slide image (WSI) classification, using multi-scale inputs to capture complementary information across different magnification levels. In such settings, high magnification provides fine-grained cellular details, while low magnification preserves global tissue-level structures, and both contribute to improved classification accuracy. Despite their effectiveness, many existing multi-stream approaches primarily emphasize the fusion of representations from different inputs rather than target-conditioned aggregation of spatial patch features. In this study, we define multiple contextual views around the ROI to capture local and broader environmental information. Our approach uses the target ROI representation to selectively aggregate spatial patch-level features from these contextual views.

\subsection{Taxonomy-Aware Supervision for Fine-Grained Recognition}

Hierarchical information has been widely used to complement flat category supervision in classification tasks~\citep{RN304,RN307,RN308}. Existing methods commonly incorporate hierarchy by modifying the finest-level classification loss or by structuring the embedding space. For example, Hierarchical Cross-Entropy (HXE) and Soft Label supervision incorporate hierarchical relationships through error penalties and target distributions, respectively~\citep{RN37}, while contrastive or metric-learning methods encourage feature distributions that reflect hierarchical relationships~\citep{RN38,RN300,RN301}. Although effective, these approaches typically inject hierarchical information into the finest-level objective through modified losses, target distributions, or sample relationships. Auxiliary classification provides another way to exploit hierarchical labels by adding supervision beyond the finest label~\citep{RN26}. Prior studies have used level-wise classifiers based on attribute hierarchy~\citep{RN27} or coarse-to-fine disease labels~\citep{RN28}, showing that auxiliary prediction tasks can help shared representations capture information across multiple levels of abstraction. Following this direction, we use taxonomy as auxiliary supervision for ROI-guided fine-grained marine organism recognition. Specifically, level-wise taxonomic cross-entropy (CE) supervision employs separate auxiliary classifiers to predict labels at higher taxonomic ranks during training, encouraging the shared representation to encode hierarchical information.

\section{Method}

\subsection{Problem Definition}

We address ROI-guided fine-grained marine organism recognition in underwater images. 
Given an underwater image $I$ and a region of interest (ROI) $b$ indicating the target organism, the goal is to predict the target class label $y$ for the organism. 
The ROI can be provided through manual annotation or obtained from an object detection model.

Formally, each training sample is represented as
\begin{equation}
    (I_i, b_i, y_i),
\end{equation}
where $I_i$ denotes the $i$-th underwater image, $b_i$ denotes the bounding box of the target organism, and $y_i \in \mathcal{Y}$ denotes its class label. 
The objective is to learn a recognition model $F_{\theta}$ such that
\begin{equation}
    \hat{y}_i = F_{\theta}(I_i, b_i),
\end{equation}
where $\hat{y}_i$ denotes the predicted class label of the organism specified by $b_i$.

Unlike standard ROI-only classification, which uses only the cropped target region, this setting retains the surrounding image regions as potential sources of contextual information. In underwater environments, such context can include nearby organisms, substrate type, habitat structure, and other environmental cues that may be informative for fine-grained taxonomic recognition.
Therefore, the model should focus on the target organism while selectively exploiting its surrounding visual context.

To this end, MATANet constructs two types of inputs from each image--ROI pair $(I_i, b_i)$: an ROI crop that focuses on the target organism and multiple contextual views that provide surrounding environmental information. The contextual views are generated by expanding the ROI at multiple scales while preserving its center. Each contextual view contains the target organism and a progressively wider surrounding region as $s$ increases. Specifically, the ROI crop and ROI-centered contextual crops are defined as
\begin{equation}
    x_i^{r} = \mathcal{C}(I_i, b_i), \quad
    x_i^{c,s} = \mathcal{C}(I_i, \phi_s(b_i)), \quad s \in \mathcal{S},
\end{equation}
where $\mathcal{C}(I,b)$ denotes the operation that crops region $b$ from image $I$ and resizes it to the model input resolution. Here, $\phi_s(b_i)$ expands the bounding box $b_i$ by a scale factor $s$ while preserving its center, $x_i^{r}$ denotes the ROI crop, $x_i^{c,s}$ denotes the contextual crop at scale $s$, and $\mathcal{S}$ is the set of context scale factors. If the expanded region exceeds the image boundary, it is clipped to the valid image extent. These multi-scale inputs allow MATANet to focus on the target organism while incorporating contextual cues from progressively wider surrounding regions.

\subsection{Overview}

\begin{figure*}[t]
  \centering
  \includegraphics[width=1.0\linewidth]{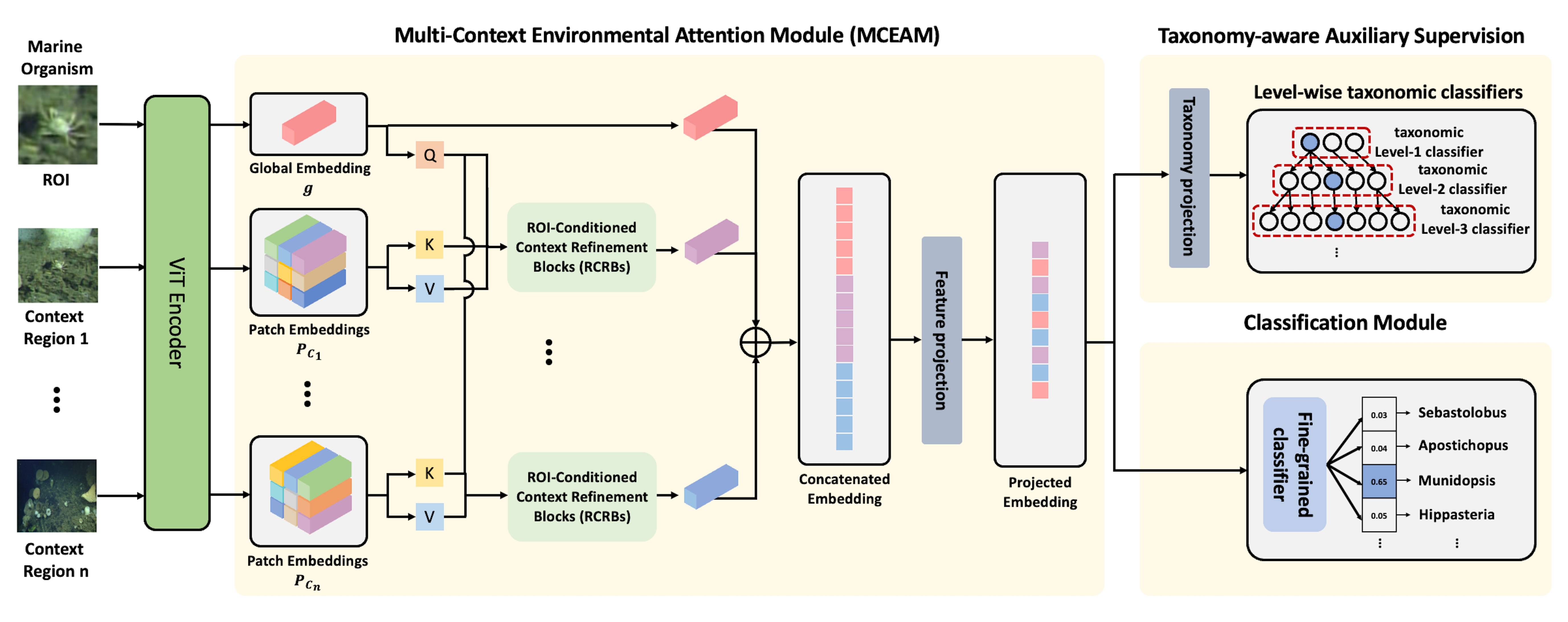}
  \caption{Overview of MATANet. A shared ViT backbone encodes the ROI crop and multi-scale contextual views. For each contextual view, MCEAM employs a stack of ROI-Conditioned Context Refinement Blocks (RCRBs), in which an ROI-conditioned query progressively aggregates relevant information from contextual patch features. The resulting context-aware representations are fused with the original ROI representation for fine-grained classification. Taxonomic information is incorporated through auxiliary taxonomy-level classifiers used only during training.}
  \label{fig:fig2}
\end{figure*}

We propose MATANet (Multi-Context Attention and Taxonomy-Aware Network), a framework for ROI-guided fine-grained marine organism recognition that combines multi-scale contextual modeling with taxonomy-aware auxiliary supervision. As illustrated in Figure~\ref{fig:fig2}, MATANet processes an ROI crop corresponding to the target organism together with multiple ROI-centered contextual views at different spatial scales. All input views are encoded using a shared Vision Transformer (ViT) backbone~\citep{RN309}. The [CLS] embedding of the ROI crop represents the target organism, whereas the patch embeddings from each contextual view retain spatially localized information about the surrounding scene.

The encoded representations are then processed by the Multi-Context Environmental Attention Module (MCEAM). For each contextual view, MCEAM applies a shared stack of ROI-Conditioned Context Refinement Blocks (RCRBs) to progressively update an ROI-conditioned query $g_c$, initialized from the ROI embedding $g$, by attending to the corresponding contextual patch embeddings $P_c$. The resulting context-aware representations are concatenated with the original ROI embedding and projected to form a fused representation $z$ for fine-grained classification.

MATANet further incorporates the taxonomic hierarchy through level-wise auxiliary supervision. Separate auxiliary classifiers predict the available higher-level taxonomic labels during training, encouraging the fused representation $z$ to encode taxonomically consistent information. The main classifier then uses this taxonomy-informed fused representation $z$ to produce the final fine-grained class prediction, whereas the auxiliary classifiers are not used during inference.

\subsection{Multi-Context Environmental Attention Module}

\begin{figure*}[t]
  \centering
  \includegraphics[width=0.3\linewidth]{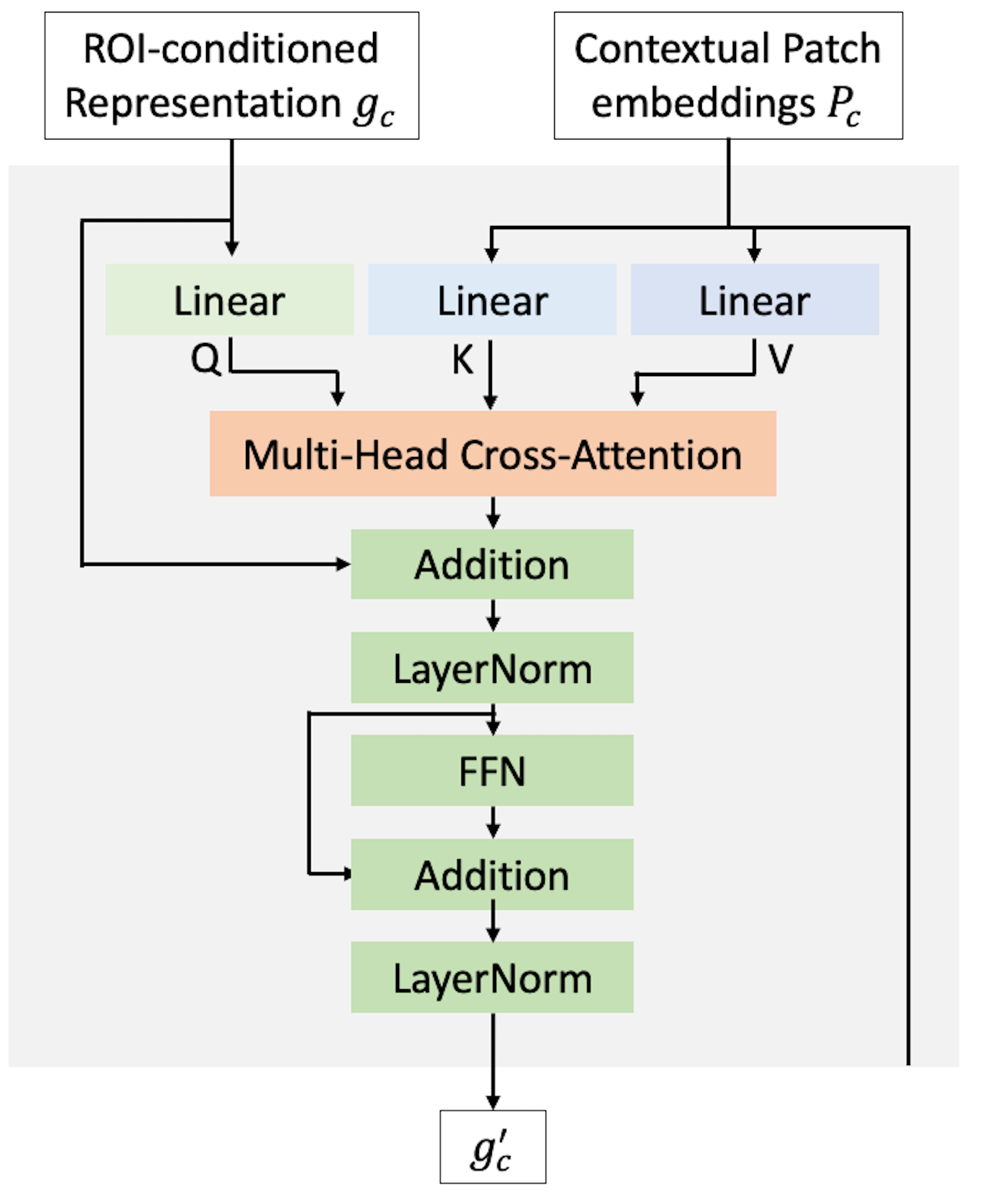}
  \caption{Architecture of the ROI-Conditioned Context Refinement Block (RCRB). The ROI-conditioned representation $g_c$ is used as the query, whereas the contextual patch embeddings $P_c$ serve as keys and values. The query is refined through multi-head cross-attention and a feed-forward network, with residual connections and layer normalization applied after each sublayer, producing the refined representation $g_c^{\prime}$.}
  \label{fig:rcrb}
\end{figure*}

Marine organisms are often closely associated with their surrounding habitats. Consequently, environmental context can provide complementary cues for fine-grained recognition beyond the visual appearance of the target organism. However, conventional ROI-only recognition primarily relies on features extracted from the target region and does not explicitly exploit contextual information relevant to the target organism specified by the ROI. To address this limitation, we propose the Multi-Context Environmental Attention Module (MCEAM), which performs ROI-conditioned aggregation of spatially localized contextual patch features from multiple contextual views.

Each input view is encoded using the shared Vision Transformer (ViT) backbone. Specifically, we extract the [CLS] embedding $g \in \mathbb{R}^{1 \times d}$ from the ROI crop and the patch embeddings
\begin{equation}
P_c = [p_c^1, p_c^2, \ldots, p_c^{n_c}]^{\top} \in \mathbb{R}^{n_c \times d}
\end{equation}
from each contextual view $c \in C$, where $C$ denotes the set of contextual views and $n_c$ is the number of patch embeddings in view $c$. The ROI embedding $g$ represents the specified target organism, whereas $P_c$ retains spatially localized information from contextual view $c$.

For each contextual view, MCEAM applies a shared stack of ROI-Conditioned Context Refinement Blocks (RCRBs). Inspired by the Transformer architecture~\citep{RN18}, each RCRB consists of a multi-head cross-attention sublayer and a feed-forward sublayer, each equipped with a residual connection and layer normalization. Through repeated application of these blocks, the ROI-conditioned representation is progressively refined by selectively aggregating target-relevant information from the contextual patch embeddings.

Given the ROI embedding $g$ and the contextual patch embeddings $P_c$ of contextual view $c$, the ROI-conditioned query representation is initialized as
\begin{equation}
g_c = g.
\end{equation}

As illustrated in Figure~\ref{fig:rcrb}, each RCRB first projects the current ROI-conditioned query representation and contextual patch embeddings into query, key, and value representations:
\begin{equation}
q_c = g_c W_Q, \quad
K_c = P_c W_K, \quad
V_c = P_c W_V,
\label{eq:rcrb_projection}
\end{equation}
where $W_Q$, $W_K$, and $W_V$ are the learnable projection matrices of the current RCRB and are shared across contextual views.

The cross-attention weights are computed as
\begin{equation}
A_c =
\mathrm{Softmax}
\left(
\frac{q_c K_c^{\top}}{\sqrt{d_h}}
\right),
\label{eq:rcrb_weights}
\end{equation}
where $d_h$ denotes the feature dimension of each attention head. The contextual information is then aggregated as
\begin{equation}
o_c = A_c V_c.
\label{eq:rcrb_aggregation}
\end{equation}
For notational simplicity, the attention-head index is omitted. In practice, the attention operation is performed in parallel over multiple heads, and the resulting head outputs are concatenated and linearly projected to obtain $o_c$.

The ROI-conditioned query representation is updated through a residual connection and layer normalization:
\begin{equation}
\bar{g}_c
=
\mathrm{LN}_{1}
\left(
g_c + o_c
\right).
\label{eq:rcrb_attention}
\end{equation}
A feed-forward network is subsequently applied with a second residual connection:
\begin{equation}
g'_c
=
\mathrm{LN}_{2}
\left(
\bar{g}_c
+
\mathrm{FFN}
\left(
\bar{g}_c
\right)
\right),
\label{eq:rcrb_ffn}
\end{equation}
where $g'_c$ denotes the refined ROI-conditioned context representation.

The RCRB output $g'_c$ contains target-relevant contextual information aggregated from the patch embeddings of contextual view $c$ under the guidance of the ROI-conditioned query. In our implementation, a stack of four RCRBs is sequentially applied to each contextual view. The RCRB stack is shared across contextual views, with separate parameters for each block. At each refinement step, the output of the preceding RCRB is used as the query representation for the subsequent block, while the same contextual patch embeddings $P_c$ are retained as keys and values. Each RCRB uses eight attention heads, and its feed-forward network consists of two fully connected layers with a ReLU activation. An attention dropout rate of 0.3 is applied. The query representation obtained from the final RCRB is denoted by $h_c$.

The context-aware representations from all contextual views are concatenated with the original ROI embedding and projected to form a unified latent representation:
\begin{equation}
z =
\mathrm{Proj}
\left(
\mathrm{Concat}
\left(
g, h_{c_1}, h_{c_2}, \ldots, h_{c_{|C|}}
\right)
\right),
\label{eq:mceam}
\end{equation}
where $(c_1,c_2,\ldots,c_{|C|})$ denotes the contextual views in a fixed scale order, and $\mathrm{Proj}(\cdot)$ denotes a learnable feature projection network.

The resulting representation $z$ combines the original ROI information with target-conditioned contextual features aggregated from the multi-scale contextual views. Through this process, MCEAM assigns different attention weights to individual contextual patch locations according to the ROI-conditioned query. These attention weights can be mapped back to the corresponding patch grid to inspect the spatial attention patterns produced by MCEAM.

\subsection{Taxonomy-Aware Auxiliary Supervision}

The taxonomic hierarchy provides structured prior knowledge beyond flat category labels for marine organism recognition. Although visual similarity does not always correspond directly to taxonomic proximity, higher taxonomic ranks encode relationships among target taxa and can provide complementary supervision for representation learning. We therefore incorporate biological taxonomy through level-wise auxiliary classification to encourage taxonomically structured representation learning.

Given the fused representation $z$ produced by MCEAM, a taxonomy projection $f_{\mathrm{tax}}(\cdot)$ generates an auxiliary representation
\begin{equation}
\tilde{z} = f_{\mathrm{tax}}(z).
\end{equation}
This projection provides a separate pathway for auxiliary taxonomic prediction while allowing its gradients to guide the fused representation $z$. The auxiliary representation $\tilde{z}$ is provided to a set of level-specific classifiers $f_{\ell}^{\mathrm{aux}}(\cdot)$, where each classifier corresponds to a higher taxonomic rank $\ell \in \mathcal{H}$, such as order, family, or genus.

Because taxonomic labels may be missing or may correspond to the finest target category of a sample, each auxiliary classifier is trained only with samples that have a valid higher-level label at the corresponding rank. Let $\mathcal{B}_{\ell}$ denote the set of eligible samples for taxonomic rank $\ell$. The auxiliary loss at rank $\ell$ is defined as
\begin{equation}
\mathcal{L}_{\ell}
=
\frac{1}{|\mathcal{B}_{\ell}|}
\sum_{i \in \mathcal{B}_{\ell}}
\mathrm{CE}
\left(
f_{\ell}^{\mathrm{aux}}(\tilde{z}_{i}),
y_{i}^{\ell}
\right),
\label{eq:rank_loss}
\end{equation}
where $y_{i}^{\ell}$ denotes the ground-truth label of sample $i$ at rank $\ell$.

The level-wise taxonomic cross-entropy loss is obtained by averaging the auxiliary losses over the valid taxonomic ranks:
\begin{equation}
\mathcal{L}_{\mathrm{lw-tax}}
=
\frac{1}{|\mathcal{H}_{\mathrm{valid}}|}
\sum_{\ell \in \mathcal{H}_{\mathrm{valid}}}
\mathcal{L}_{\ell},
\label{eq:tax_loss}
\end{equation}
where $\mathcal{H}_{\mathrm{valid}}$ denotes the set of taxonomic ranks with at least one eligible sample in the current mini-batch. Thus, the main classifier is trained using the finest available category of each sample, whereas the auxiliary classifiers are trained only on its valid higher-level taxonomic ancestors. Samples with missing labels or without a higher-level label at a given rank are excluded from the corresponding auxiliary loss. This avoids duplicate supervision at the final prediction level while accommodating incomplete and mixed-depth taxonomic annotations.

\subsection{Objective Function}

The main classifier $f_{\mathrm{cls}}(\cdot)$ takes the fused representation $z$ produced by MCEAM and predicts the finest available category assigned to each sample. For a mini-batch $\mathcal{B}$, the main classification loss is defined as
\begin{equation}
\mathcal{L}_{\mathrm{cls}}
=
\frac{1}{|\mathcal{B}|}
\sum_{i \in \mathcal{B}}
\mathrm{CE}
\left(
f_{\mathrm{cls}}(z_i),
y_i^{*}
\right),
\label{eq:cls}
\end{equation}
where $y_i^{*}$ denotes the finest available ground-truth label of sample $i$.

The overall training objective of MATANet is defined as
\begin{equation}
\mathcal{L}_{\mathrm{total}}
=
\mathcal{L}_{\mathrm{cls}}
+
\lambda \mathcal{L}_{\mathrm{lw-tax}},
\label{eq:total}
\end{equation}
where $\lambda$ controls the contribution of the level-wise taxonomic cross-entropy loss. Unless otherwise specified, $\lambda$ is set to 1 in all experiments. The overall objective optimizes fine-grained classification while using higher-rank taxonomic supervision to encourage taxonomically consistent representations.

\section{Experiments}

\subsection{Datasets}

\begin{figure*}[t]
  \centering
   \includegraphics[width=1.0\linewidth]{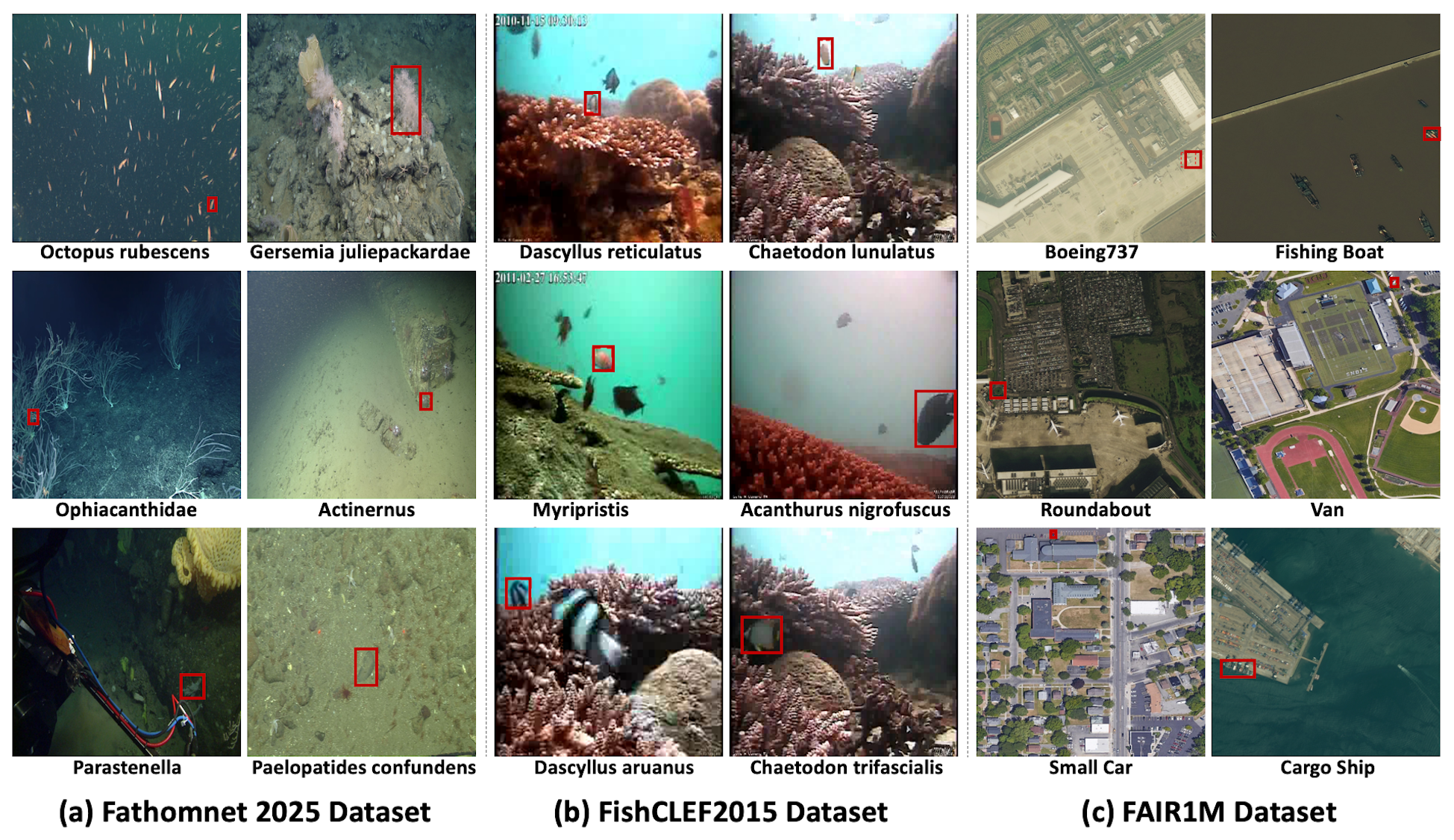}
    \caption{Example images from the FathomNet 2025 \citep{RN19}, FishCLEF2015 \citep{RN21}, and FAIR1M \citep{RN20} datasets, illustrating differences in image quality, target appearance, and surrounding environmental context.}
   \label{fig:dataset}
\end{figure*}

We selected datasets that allow the proposed ROI-guided recognition framework to be evaluated using object-level ROI annotations and hierarchical label information. Because MATANet constructs ROI-centered multi-scale contextual regions and uses hierarchy-aware auxiliary supervision, the evaluation requires object-level ROI annotations and a usable hierarchical label structure. Many fish or underwater organism datasets \citep{RN11,RN109,RN110,RN111} provide only image-level labels or cropped object images without bounding box annotations, making it difficult to construct ROI-centered contextual regions. Accordingly, we use FathomNet 2025 \citep{RN19} and FishCLEF2015 \citep{RN21} as marine organism datasets that satisfy these requirements. We additionally include FAIR1M \citep{RN20}, a high-resolution satellite imagery dataset, as a supplementary non-underwater benchmark to examine the applicability of the proposed framework beyond marine imagery. Figure~\ref{fig:dataset} shows example images from the three datasets, illustrating differences in image quality, target appearance, and surrounding environmental context.

FathomNet 2025 is a high-resolution underwater image subset of the FathomNet database \citep{RN29}. It contains annotations for 79 marine organism categories spanning seven biological phyla across diverse habitats. The annotations provide bounding box-level ROIs together with the finest available taxonomic labels. The training set contains 8,981 images and 23,700 ROIs, while the official test set includes 325 images and 788 ROIs. Official test-set evaluation is conducted through the challenge evaluation system, which provides standardized performance results for the held-out test set \citep{RN19}. For ablation analysis, we constructed a fixed 80\%/20\% image-level stratified training/validation split using a fixed random seed. The same split was used for all configurations and repeated runs.

FishCLEF2015 is a dataset derived from underwater videos with a resolution of 320$\times$240, collected through the Fish4Knowledge project. It provides annotations for 15 fish categories across five taxonomic orders inhabiting near-shore coral reef environments. Only samples with ROI annotations are used for both training and evaluation. The training set comprises 5,278 images with 9,160 ROIs, and the test set includes 9,339 images with 13,585 ROIs.

FAIR1M v2.0 is a high-resolution satellite image dataset designed for fine-grained object detection. It provides a two-level coarse-to-fine semantic label structure, consisting of five coarse-grained superclasses, such as aircraft, ship, and vehicle, and 37 fine-grained subclasses. The training set contains 16,488 images with 396,615 ROIs, while the validation set contains 8,287 images with 202,957 ROIs and is used as the test set in our experiments.

For FathomNet 2025 and FishCLEF2015, taxonomic paths were obtained using the fathomnet.api taxonomy information. The main classifier predicts the finest available category provided for each sample, even when the finest categories correspond to different taxonomic ranks. The auxiliary classifiers use valid ancestral labels among phylum, class, order, family, and genus, while the finest target label itself is excluded from auxiliary supervision. For example, when the finest target label is a family, the family label is used only by the main classifier and is not used as an auxiliary target. If an intermediate taxonomic rank is unavailable, the sample is excluded only from the auxiliary loss at that rank. For FAIR1M, the coarse superclass is used as the single auxiliary hierarchical target. The same FathomNet hierarchy is used for both the internal validation evaluation and the official test evaluation.

In summary, the three datasets provide complementary evaluation settings. FathomNet 2025 serves as the primary benchmark because it contains high-resolution underwater images, diverse habitats, object-level ROIs, and a relatively rich taxonomic hierarchy. FishCLEF2015 provides a more constrained underwater setting with lower-resolution images, fewer categories, and a simpler taxonomic structure. FAIR1M provides a supplementary evaluation setting with a shallow two-level semantic hierarchy and substantially different target--context relationships from those observed in underwater imagery. This allows us to examine the proposed design under limited hierarchical depth and in a distinct visual domain.

\subsection{Evaluation Metrics}

We evaluate performance using Accuracy (ACC), macro-precision, macro-F1 score, and Hierarchical Distance (HD). 
ACC measures the proportion of correctly classified samples, whereas macro-averaged metrics give equal weight to each class. 
Higher ACC, macro-precision, and macro-F1 score indicate better flat classification performance, while lower HD indicates better hierarchical consistency.

Given $N$ evaluation samples, ACC is defined as
\begin{equation}
\mathrm{ACC} = \frac{1}{N} \sum_{i=1}^{N} \mathbb{I}(\hat{y}_i = y_i^{*}),
\label{eq:acc}
\end{equation}
where $y_i^{*}$ and $\hat{y}_i$ denote the ground-truth finest label and predicted label, respectively.

For each class $k$, Precision, Recall, and F1-score are computed as
\begin{equation}
\mathrm{Precision}_k = \frac{TP_k}{TP_k + FP_k}, \quad
\mathrm{Recall}_k = \frac{TP_k}{TP_k + FN_k},
\label{eq:precision_recall}
\end{equation}
\begin{equation}
\mathrm{F1}_k =
\frac{2 \cdot \mathrm{Precision}_k \cdot \mathrm{Recall}_k}
{\mathrm{Precision}_k + \mathrm{Recall}_k},
\label{eq:f1_class}
\end{equation}
where $TP_k$, $FP_k$, and $FN_k$ denote the true positives, false positives, and false negatives for class $k$. 
When the denominator of a class-wise metric is zero, the corresponding value is set to zero. 
Macro-precision and macro-F1 score are defined as
\begin{equation}
\mathrm{Precision}_{\mathrm{macro}} =
\frac{1}{K} \sum_{k=1}^{K} \mathrm{Precision}_k, \quad
\mathrm{F1}_{\mathrm{macro}} =
\frac{1}{K} \sum_{k=1}^{K} \mathrm{F1}_k.
\label{eq:macro_metrics}
\end{equation}

where $K$ denotes the number of classes considered in the evaluation.
HD measures the shortest-path distance between the predicted and ground-truth labels in a hierarchy tree. 
Unlike flat classification metrics, HD reflects whether an incorrect prediction remains close to the ground-truth label in the hierarchy. 
Thus, HD is used to evaluate the hierarchical consistency of model predictions. 
Assuming a unit distance between adjacent nodes, HD is computed as
\begin{equation}
\mathrm{HD} = \frac{1}{N} \sum_{i=1}^{N} d(y_i^{*}, \hat{y}_i),
\label{eq:HD}
\end{equation}
where $d(\cdot)$ denotes the shortest-path distance between two nodes in the hierarchy.

For FathomNet 2025, official test-set performance is obtained through the challenge evaluation system and reported using HD \citep{RN19}. 
For ablation analysis on FathomNet 2025, we use the image-level training/validation split described in the dataset section and report ACC, macro-precision, macro-F1 score, and HD. For FishCLEF2015 and FAIR1M, we report the same four metrics.

\subsection{Implementation Details}
\label{imp}

All experiments were conducted on an Ubuntu 22.04 system equipped with an Intel Core i9-10980XE CPU and an NVIDIA RTX A6000 GPU, using Python 3.10 and PyTorch 2.6.0. Object regions were cropped using the bounding box annotations and resized to 256$\times$256. Contextual inputs were constructed as square ROI-centered crops at scales of 3$\times$ and 5$\times$ relative to the bounding box, together with the full image, and were resized to the same resolution. For FAIR1M, axis-aligned ROI crops were generated from the center, width, and height of the oriented bounding boxes without rotation. The augmentation pipeline consisted of random horizontal and vertical flipping, color jitter, and rotation. The same augmentation types were applied to the ROI and contextual views, with transformation parameters sampled independently for each view.

Unless otherwise specified, MATANet refers to the M4 configuration using the official DINOv2-pretrained ViT-B/14 backbone~\citep{RN33}. M4-S and M4-L use the same architecture with ViT-S/14 and ViT-L/14 backbones, respectively. All backbones were fine-tuned during training. MCEAM used four cross-attention blocks with eight attention heads. The feature projection, main classification head, and auxiliary taxonomic classifiers were implemented using fully connected layers. The weight of each auxiliary hierarchy-aware loss was set to $\lambda=1$. All MATANet configurations were optimized using AdamW with a weight decay of 0.001. For FathomNet 2025 and FAIR1M, the learning rates were set to $1\times10^{-6}$ for the backbone and $1\times10^{-5}$ for the remaining modules. For FishCLEF2015, a learning rate of $1\times10^{-5}$ was used for all modules. The models were trained for 25 epochs with a batch size of 16 on FathomNet 2025, 50 epochs with a batch size of 64 on FishCLEF2015, and 25 epochs with a batch size of 64 on FAIR1M. The checkpoint from the final epoch was used for evaluation.

Publicly available implementations were used for the benchmark methods whenever possible. For INTR~\citep{RN35}, TransFG~\citep{RN36}, and MSDBN~\citep{RN15}, we followed their architecture-specific preprocessing and default training settings. The remaining benchmark models were trained using the same dataset splits, input resolution, augmentation pipeline, and optimization protocol as MATANet. All experiments were repeated three times with different random seeds, and the mean and standard deviation are reported.

\subsection{Comparison with Benchmark Models}
\begin{table}[t]
\centering
\small
\setlength{\tabcolsep}{4pt}
\renewcommand{\arraystretch}{1.05}
\begin{tabular}{lcc}
\toprule
\multirow{2}{*}{\textbf{Method}} & \multicolumn{2}{c}{\textbf{Hierarchical Distance (HD)} $\downarrow$} \\
\cmidrule(lr){2-3}
& \textbf{Public} & \textbf{Private} \\
\midrule
VGG19~\citep{RN30} & 4.993 $\pm$ 0.186 & 5.050 $\pm$ 0.151 \\
ResNet50~\citep{RN31} & 3.967 $\pm$ 0.115 & 3.863 $\pm$ 0.086 \\
ResNeXt50~\citep{RN32} & 4.107 $\pm$ 0.186 & 3.880 $\pm$ 0.225 \\
Swin-B~\citep{RN017} & 3.023 $\pm$ 0.129 & 2.650 $\pm$ 0.087 \\
MaxViT-T~\citep{RN34} & 3.640 $\pm$ 0.147 & 3.597 $\pm$ 0.176 \\
DINOv2-B~\citep{RN33} & 3.160 $\pm$ 0.101 & 2.777 $\pm$ 0.050 \\
TransFG~\citep{RN36} & 2.890 $\pm$ 0.221 & 2.620 $\pm$ 0.105 \\
INTR~\citep{RN35} & 3.973 $\pm$ 0.287 & 3.773 $\pm$ 0.335 \\
MSDBN~\citep{RN15} & \underline{2.763} $\pm$ 0.130 & \underline{2.603} $\pm$ 0.076 \\
MATANet (M4) & \textbf{1.740} $\pm$ 0.030 & \textbf{1.570} $\pm$ 0.020 \\
\bottomrule
\end{tabular}

\caption{Comparison of benchmark models on the FathomNet 2025 dataset using hierarchical distance (HD). Models were evaluated on the Public and Private test sets through the official challenge evaluation system. The best and second-best results were determined using the unrounded values and are shown in bold and underlined, respectively.}
\label{table:t1}
\end{table}

Previous studies on fine-grained marine organism recognition have faced challenges in ensuring reproducibility and fair benchmarking, as source code is often not publicly available and the datasets used in each study vary substantially. Recently, Fish-Vista \citep{RN11} evaluated multiple benchmark models under a consistent experimental setting using a curated dataset mainly composed of fish categories. In this study, we adopt a similar benchmarking strategy and compare MATANet with widely used baseline models whose implementations are publicly available. The comparison includes CNN-based models, such as VGG19 \citep{RN30}, ResNet50 \citep{RN31}, and ResNeXt50 \citep{RN32}; transformer-based vision models, such as Swin Transformer \citep{RN017}, DINOv2 \citep{RN33}, and MaxViT \citep{RN34}; and FGVC models, such as TransFG \citep{RN36} and INTR \citep{RN35}. In addition, we include MSDBN \citep{RN15}, a fish-specific FGVC model with a publicly available implementation, as an additional comparison method. Unless otherwise specified, MATANet refers to the M4 configuration selected from the context ablation study (see Table~\ref{table:context}). Detailed implementation settings are provided in Section~\ref{imp}.

To assess the effectiveness of MATANet, we first compare it with representative benchmark models using the official FathomNet 2025 evaluation system. Table~\ref{table:t1} reports the hierarchical distance (HD) of MATANet (M4) and the benchmark models on the Public and Private test splits. MATANet achieves the lowest mean HD on both splits, with values of 1.740 and 1.570 on the Public and Private splits, respectively. Among the benchmark models, MSDBN achieves the second-lowest mean HD on both splits, indicating that a fish-specific FGVC architecture remains competitive in this broader marine organism recognition setting. In general, conventional CNN-based models yield higher HD values, whereas transformer-based and FGVC-specific models, including Swin Transformer, DINOv2-B, TransFG, and MSDBN, provide stronger baselines. Importantly, the DINOv2-B baseline corresponds to the M0 configuration presented in Table~\ref{table:context} and uses the same DINOv2-pretrained ViT-B/14 backbone as M4. The improvement over this baseline indicates that the performance gain cannot be attributed solely to the backbone architecture or pretrained initialization.

We additionally report the M4-L configuration in Table~\ref{table:backbone_scaling}. Because M4-L is based on the larger DINOv2-pretrained ViT-L/14 backbone, it is excluded from the backbone-controlled comparison in Table~\ref{table:t1} and analyzed separately as a scaling experiment. M4-L achieves mean HD values of $1.717 \pm 0.131$ and $1.423 \pm 0.051$ on the Public and Private splits, respectively. Compared with M4, the improvement is modest on the Public split but more pronounced on the hidden Private split, suggesting that the proposed multi-context and taxonomy-aware design can further benefit from increased backbone capacity. Notably, the performance of M4-L is comparable to the leading results reported on the official challenge leaderboard.

\begin{table}[t]
\centering
\small
\setlength{\tabcolsep}{3pt}
\renewcommand{\arraystretch}{1.05}
\begin{tabular}{lcccc}
\toprule
\textbf{Method} & \textbf{Acc.} $\uparrow$ & \textbf{Macro-Prec.} $\uparrow$ & \textbf{Macro-F1} $\uparrow$ & \textbf{HD} $\downarrow$ \\
\midrule
VGG19~\citep{RN30}      & 0.705 $\pm$ 0.002 & 0.392 $\pm$ 0.012 & 0.346 $\pm$ 0.008 & 1.715 $\pm$ 0.014 \\
ResNet50~\citep{RN31}   & 0.727 $\pm$ 0.007 & 0.441 $\pm$ 0.008 & 0.367 $\pm$ 0.009 & 1.576 $\pm$ 0.050 \\
ResNeXt50~\citep{RN32}  & 0.700 $\pm$ 0.004 & 0.409 $\pm$ 0.005 & 0.356 $\pm$ 0.008 & 1.742 $\pm$ 0.033 \\
Swin-B~\citep{RN017}    & 0.764 $\pm$ 0.006 & 0.513 $\pm$ 0.033 & 0.426 $\pm$ 0.009 & 1.331 $\pm$ 0.034 \\
MaxViT-T~\citep{RN34}   & 0.721 $\pm$ 0.013 & 0.447 $\pm$ 0.015 & 0.387 $\pm$ 0.012 & 1.593 $\pm$ 0.102 \\
DINOv2-B~\citep{RN33}   & \underline{0.766} $\pm$ 0.010 & \underline{0.528} $\pm$ 0.022 & \underline{0.426} $\pm$ 0.018 & \underline{1.327} $\pm$ 0.059 \\
TransFG~\citep{RN36}    & 0.722 $\pm$ 0.015 & 0.439 $\pm$ 0.023 & 0.359 $\pm$ 0.009 & 1.558 $\pm$ 0.151 \\
INTR~\citep{RN35}       & 0.648 $\pm$ 0.006 & 0.412 $\pm$ 0.010 & 0.318 $\pm$ 0.001 & 1.882 $\pm$ 0.043 \\
MSDBN~\citep{RN15}      & 0.700 $\pm$ 0.010 & 0.394 $\pm$ 0.026 & 0.330 $\pm$ 0.018 & 1.684 $\pm$ 0.031 \\
MATANet (M4) & \textbf{0.793} $\pm$ 0.010 & \textbf{0.555} $\pm$ 0.012 & \textbf{0.455} $\pm$ 0.015 & \textbf{1.120} $\pm$ 0.086 \\
\bottomrule
\end{tabular}
\caption{Classification results on the FishCLEF2015 dataset evaluated using accuracy, macro-precision, macro-F1 score, and hierarchical distance (HD). The best and second-best results were determined using the unrounded values and are shown in bold and underlined, respectively.}
\label{table:fishclef}
\end{table}

To further evaluate the effectiveness of MATANet across marine organism datasets, we conduct experiments on FishCLEF2015. Table~\ref{table:fishclef} reports the performance of MATANet and the benchmark models in terms of accuracy, macro-precision, macro-F1 score, and hierarchical distance (HD). MATANet achieves the best performance across all evaluation metrics, while DINOv2-B provides the strongest overall performance among the benchmark models. In terms of HD, the performance gap between MATANet and the strongest benchmark model is smaller on FishCLEF2015 than on FathomNet 2025. This difference may partly reflect the visual and taxonomic characteristics of FishCLEF2015. First, FishCLEF2015 contains low-resolution images of 320$\times$240 and exhibits relatively limited background diversity, which may reduce the benefit of incorporating multi-scale contextual views. Second, the dataset contains fewer categories and a simpler taxonomic structure than FathomNet 2025, potentially limiting the contribution of taxonomy-aware auxiliary supervision. These observations suggest that the benefits of multi-context modeling and taxonomy-aware supervision may vary with the visual complexity and hierarchical structure of the dataset.

\subsection{Ablation Study}
\begin{table}[t]
\centering
\small
\setlength{\tabcolsep}{4pt}
\renewcommand{\arraystretch}{1.05}
\begin{tabular}{llccccc}
\toprule
\textbf{Model} & \textbf{Context} & \textbf{LW-CE} & \textbf{Acc.} $\uparrow$ & \textbf{Macro-Prec.} $\uparrow$ & \textbf{Macro-F1} $\uparrow$ & \textbf{HD} $\downarrow$ \\
\midrule
M0 & ROI only & -- & 0.846 $\pm$ 0.003 & 0.842 $\pm$ 0.003 & 0.842 $\pm$ 0.003 & 0.758 $\pm$ 0.013 \\
M1 & 3$\times$ & -- & 0.868 $\pm$ 0.005 & 0.866 $\pm$ 0.006 & 0.866 $\pm$ 0.001 & 0.608 $\pm$ 0.024 \\
M2 & 3$\times$, 5$\times$ & -- & 0.875 $\pm$ 0.002 & 0.872 $\pm$ 0.001 & 0.872 $\pm$ 0.002 & 0.550 $\pm$ 0.006 \\
M3 & 3$\times$, 5$\times$, Full & -- & \underline{0.878} $\pm$ 0.003 & \underline{0.874} $\pm$ 0.005 & \underline{0.875} $\pm$ 0.005 & \underline{0.530} $\pm$ 0.025 \\
M4 & 3$\times$, 5$\times$, Full & \checkmark & \textbf{0.881} $\pm$ 0.003 & \textbf{0.878} $\pm$ 0.003 & \textbf{0.878} $\pm$ 0.003 & \textbf{0.501} $\pm$ 0.008 \\
\bottomrule
\end{tabular}
\caption{Component ablation study on the FathomNet 2025 validation set. The 3$\times$ and 5$\times$ settings denote ROI-centered contextual crops, while Full denotes the full-image context view. LW-CE denotes the proposed level-wise taxonomic cross-entropy supervision. The best and second-best results are shown in bold and underlined, respectively.}
\label{table:context}
\end{table}

We conduct ablation studies to evaluate the contributions of multi-scale contextual views and taxonomy-aware auxiliary supervision in MATANet. Table~\ref{table:context} presents the results for different contextual-view configurations and level-wise taxonomic CE on the FathomNet 2025 validation set. The main component analysis is conducted using M0--M4 with the same ViT-B/14 backbone. Compared with the ROI-only baseline (M0), all variants incorporating additional ROI-centered contextual views improve recognition performance and reduce HD. Adding the 3$\times$ view in M1 provides a clear improvement, while incorporating the 5$\times$ and full-image views in M2 and M3 leads to further gains. These results demonstrate the benefit of progressively incorporating contextual views at broader spatial scales. Under the same contextual-view configuration as M3, adding level-wise taxonomic CE in M4 yields modest improvements in the flat classification metrics and a clearer reduction in HD. This result suggests that taxonomy-aware auxiliary supervision contributes more strongly to hierarchical consistency than to flat classification performance.

\begin{table}[t]
\centering
\small
\setlength{\tabcolsep}{3.5pt}
\renewcommand{\arraystretch}{1.05}
\begin{tabular}{lllcccc}
\toprule
\textbf{Agg.} & \textbf{Target} & \textbf{Bkg.} & \textbf{Acc.} $\uparrow$ & \textbf{Macro-Prec.} $\uparrow$ & \textbf{Macro-F1} $\uparrow$ & \textbf{HD} $\downarrow$ \\
\midrule
MCEAM  & \checkmark & --         & 0.850 $\pm$ 0.001 & 0.847 $\pm$ 0.003 & 0.847 $\pm$ 0.003 & 0.695 $\pm$ 0.019 \\
MCEAM  & --         & \checkmark & 0.869 $\pm$ 0.003 & 0.865 $\pm$ 0.005 & 0.865 $\pm$ 0.004 & 0.590 $\pm$ 0.020 \\
Concat. & \checkmark & \checkmark & 0.869 $\pm$ 0.004 & 0.866 $\pm$ 0.004 & 0.865 $\pm$ 0.004 & 0.568 $\pm$ 0.026 \\
MCEAM  & \checkmark & \checkmark & \textbf{0.881} $\pm$ 0.003 & \textbf{0.878} $\pm$ 0.003 & \textbf{0.878} $\pm$ 0.003 & \textbf{0.501} $\pm$ 0.008 \\
\bottomrule
\end{tabular}
\caption{Ablation of contextual content and aggregation strategy based on M4 on the FathomNet 2025 validation set. Agg. denotes the aggregation strategy, and Concat. denotes direct concatenation of the independently encoded ROI and contextual embeddings. Target and Bkg. indicate whether the target region and surrounding background are retained in the contextual views, respectively. When Target is not retained, pixels inside the target bounding box are masked; when Bkg. is not retained, pixels outside the target bounding box are masked. The original target remains available through the ROI branch in all variants. The best mean results are shown in bold.}
\label{table:context_fusion}
\end{table}

We conduct additional contextual-content and aggregation ablations based on M4, as reported in Table~\ref{table:context_fusion}. When the surrounding background is removed from the contextual views and only the target region is retained, accuracy decreases from 0.881 to 0.850, while HD increases from 0.501 to 0.695. When the target is masked in the contextual views while the original target remains available through the ROI branch, accuracy decreases to 0.869 and HD increases to 0.590. The performance degradation in both settings indicates that repeated observations of the target at different relative scales and information from the surrounding scene both contribute to the performance of M4. However, removing the surrounding background causes a larger degradation than removing the target from the contextual views. This result suggests that the improvement of M4 over the ROI-only baseline cannot be explained solely by repeated observations of the target at different relative scales and that the surrounding scene provides particularly important complementary evidence for recognition.

Replacing MCEAM with direct concatenation of the independently encoded ROI and contextual embeddings yields an accuracy of 0.869 and an HD of 0.568, compared with 0.881 and 0.501 for M4, respectively. The relatively strong performance of the concatenation baseline indicates that access to multiple contextual views accounts for a substantial portion of the improvement over the ROI-only baseline. Nevertheless, under the same input views, backbone, and supervision configuration, MCEAM achieves a 0.012 higher mean accuracy and a 0.067 lower mean HD than direct concatenation. MCEAM also achieves higher mean macro-precision and macro-F1 score. Beyond these numerical improvements, MCEAM performs ROI-conditioned aggregation over spatial contextual patches rather than directly combining global view-level embeddings. This formulation explicitly models target--context interactions and allows the spatial aggregation behavior to be examined through cross-attention maps.

\begin{table*}[t]
\centering
\small
\setlength{\tabcolsep}{6pt}
\renewcommand{\arraystretch}{1.05}
\begin{tabular}{llccccc}
\toprule
\textbf{Model} & \textbf{Backbone} & \textbf{Params (M)} & \textbf{FLOPs (G)} & \textbf{Time/Obj. (s)} & \textbf{Acc.} $\uparrow$ & \textbf{HD} $\downarrow$ \\
\midrule
M4-S & ViT-S/14 & 39.70 & 60.40 & 0.012 & 0.841 $\pm$ 0.004 & 0.753 $\pm$ 0.039 \\
M4 & ViT-B/14 & 156.96 & 240.49 & 0.024 & 0.881 $\pm$ 0.003 & 0.501 $\pm$ 0.008 \\
M4-L & ViT-L/14 & 429.40 & 819.64 & 0.070 & 0.955 $\pm$ 0.001 & 0.169 $\pm$ 0.006 \\
\bottomrule
\end{tabular}
\caption{Effect of backbone scaling on the FathomNet 2025 validation set. M4-S, M4, and M4-L use the same multi-context and taxonomy-aware configuration with DINOv2-pretrained ViT-S/14, ViT-B/14, and ViT-L/14 backbones, respectively. Params and FLOPs denote the number of inference-time parameters and floating-point operations per object, respectively. Time/Obj. denotes the average inference time per object with a batch size of 1.}
\label{table:backbone_scaling}
\end{table*}

Table~\ref{table:backbone_scaling} examines the effect of backbone scaling on the performance and computational cost of MATANet. Replacing the ViT-S/14 backbone with ViT-B/14 improves accuracy from 0.841 to 0.881 and reduces HD from 0.753 to 0.501. Further scaling to ViT-L/14 yields an accuracy of 0.955 and an HD of 0.169, showing that the proposed multi-context and taxonomy-aware configuration benefits substantially from increased backbone capacity. These gains, however, are accompanied by higher computational costs. The number of parameters increases from 39.70M for M4-S to 156.96M for M4 and 429.40M for M4-L, while the inference time increases from 0.012 to 0.024 and 0.070 seconds per object, respectively. Overall, the results indicate a clear trade-off between recognition performance and computational efficiency across backbone scales.

\begin{table}[t]
\centering
\small
\setlength{\tabcolsep}{4pt}
\renewcommand{\arraystretch}{1.05}
\begin{tabular}{lcccc}
\toprule
\textbf{Aux. Ranks} & \textbf{Acc.} $\uparrow$ & \textbf{Macro-Prec.} $\uparrow$ & \textbf{Macro-F1} $\uparrow$ & \textbf{HD} $\downarrow$ \\
\midrule
None~(M3)   & 0.878 $\pm$ 0.003 & 0.874 $\pm$ 0.005 & 0.875 $\pm$ 0.005 & 0.530 $\pm$ 0.025 \\
Rank $\leq$ 1    & 0.876 $\pm$ 0.004 & 0.873 $\pm$ 0.003 & 0.872 $\pm$ 0.003 & 0.537 $\pm$ 0.028 \\
Rank $\leq$ 2    & 0.878 $\pm$ 0.003 & 0.875 $\pm$ 0.003 & 0.875 $\pm$ 0.003 & 0.524 $\pm$ 0.008 \\
Rank $\leq$ 3    & 0.879 $\pm$ 0.004 & \underline{0.876} $\pm$ 0.004 & 0.876 $\pm$ 0.004 & 0.520 $\pm$ 0.014 \\
Rank $\leq$ 4    & \underline{0.880} $\pm$ 0.004 & 0.875 $\pm$ 0.004 & \underline{0.876} $\pm$ 0.004 & \underline{0.515} $\pm$ 0.023 \\
Rank $\leq$ 5~(M4)   & \textbf{0.881} $\pm$ 0.003 & \textbf{0.878} $\pm$ 0.003 & \textbf{0.878} $\pm$ 0.003 & \textbf{0.501} $\pm$ 0.008 \\
\bottomrule
\end{tabular}

\caption{Ablation study of auxiliary taxonomic ranks used in level-wise taxonomic CE on the FathomNet 2025 validation set. All variants use the same configuration as M3 and differ only in the auxiliary taxonomic ranks. Rank indices 1--5 correspond to phylum, class, order, family, and genus, respectively. Rank $\leq k$ denotes the cumulative use of valid auxiliary classifiers from phylum through rank index $k$. The finest available target label is excluded from auxiliary supervision. The best and second-best results were determined using the unrounded values and are shown in bold and underlined, respectively.}
\label{table:taxlevel}
\end{table}

Table~\ref{table:taxlevel} shows that using only phylum-level supervision does not improve performance over M3, indicating that very coarse taxonomic supervision alone provides limited guidance for fine-grained recognition. As finer taxonomic ranks are cumulatively incorporated, HD generally decreases, and the configuration using all valid ranks achieves the best mean performance across all metrics. These results suggest that supervision across multiple taxonomic ranks provides complementary hierarchical information and improves taxonomic consistency. Accordingly, all valid auxiliary ranks are used in M4.

\begin{table}[t]
\centering
\small
\setlength{\tabcolsep}{4pt}
\renewcommand{\arraystretch}{1.05}
\begin{tabular}{llcccc}
\toprule
\textbf{Main Loss} & \textbf{Aux. Loss} & \textbf{Acc.} $\uparrow$ & \textbf{Macro-Prec.} $\uparrow$ & \textbf{Macro-F1} $\uparrow$ & \textbf{HD} $\downarrow$ \\
\midrule
CE~(M3) & -- & 0.878 $\pm$ 0.003 & 0.874 $\pm$ 0.005 & 0.875 $\pm$ 0.005 & 0.530 $\pm$ 0.025 \\
SoftLabel & -- & 0.871 $\pm$ 0.001 & 0.866 $\pm$ 0.001 & 0.866 $\pm$ 0.001 & 0.535 $\pm$ 0.011 \\
HXE & -- & 0.871 $\pm$ 0.001 & 0.869 $\pm$ 0.001 & 0.868 $\pm$ 0.001 & 0.537 $\pm$ 0.009 \\
CE & HMCE & 0.875 $\pm$ 0.002 & 0.871 $\pm$ 0.002 & 0.871 $\pm$ 0.002 & 0.560 $\pm$ 0.015 \\
CE & SoftLabel & 0.881 $\pm$ 0.002 & 0.877 $\pm$ 0.003 & 0.878 $\pm$ 0.003 & 0.526 $\pm$ 0.007 \\
CE & HXE & \textbf{0.883} $\pm$ 0.001 & \textbf{0.880} $\pm$ 0.002 & \textbf{0.880} $\pm$ 0.001 & \underline{0.514} $\pm$ 0.006 \\
CE~(M4) & LW-CE & \underline{0.881} $\pm$ 0.003 & \underline{0.878} $\pm$ 0.003 & \underline{0.878} $\pm$ 0.003 & \textbf{0.501} $\pm$ 0.008 \\
\bottomrule
\end{tabular}
\caption{Comparison of hierarchy-aware learning strategies on the FathomNet 2025 validation set. All variants use the same multi-scale context configuration as M3 and differ only in the main loss or auxiliary supervision strategy. The first three rows compare main classification losses without auxiliary supervision, whereas the next three rows use CE as the main loss with different auxiliary objectives. The final row corresponds to M4 with the proposed level-wise cross-entropy (LW-CE) auxiliary supervision. CE denotes cross-entropy. The best and second-best results were determined using the unrounded values and are shown in bold and underlined, respectively.}
\label{table:hierloss}
\end{table}

Table~\ref{table:hierloss} further examines alternative approaches for incorporating hierarchical information into training. SoftLabel and HXE are implemented following \citet{RN37}, whereas HMCE follows \citet{RN38}. Using M3 as the base configuration, we compare two types of hierarchy-aware strategies: replacing the main fine-grained classification loss with SoftLabel or HXE, and introducing SoftLabel, HXE, HMCE, or LW-CE as an auxiliary training objective. Replacing the main CE loss with SoftLabel or HXE does not improve performance over the CE baseline, suggesting that directly modifying the finest-label objective provides limited benefit in this setting. In contrast, most auxiliary variants improve either the flat classification metrics or HD, indicating that hierarchical information is more effective when used to complement, rather than replace, the main fine-grained objective. Among the auxiliary strategies, HXE achieves the best flat classification performance, whereas the proposed LW-CE achieves the lowest HD. These results demonstrate that, within the proposed multi-context aggregation framework, auxiliary hierarchy-aware supervision effectively improves hierarchical consistency while preserving strong flat classification performance.

\subsection{Qualitative and Representation Analysis}

\begin{figure*}[t]
\centering
\includegraphics[width=1.0\linewidth]{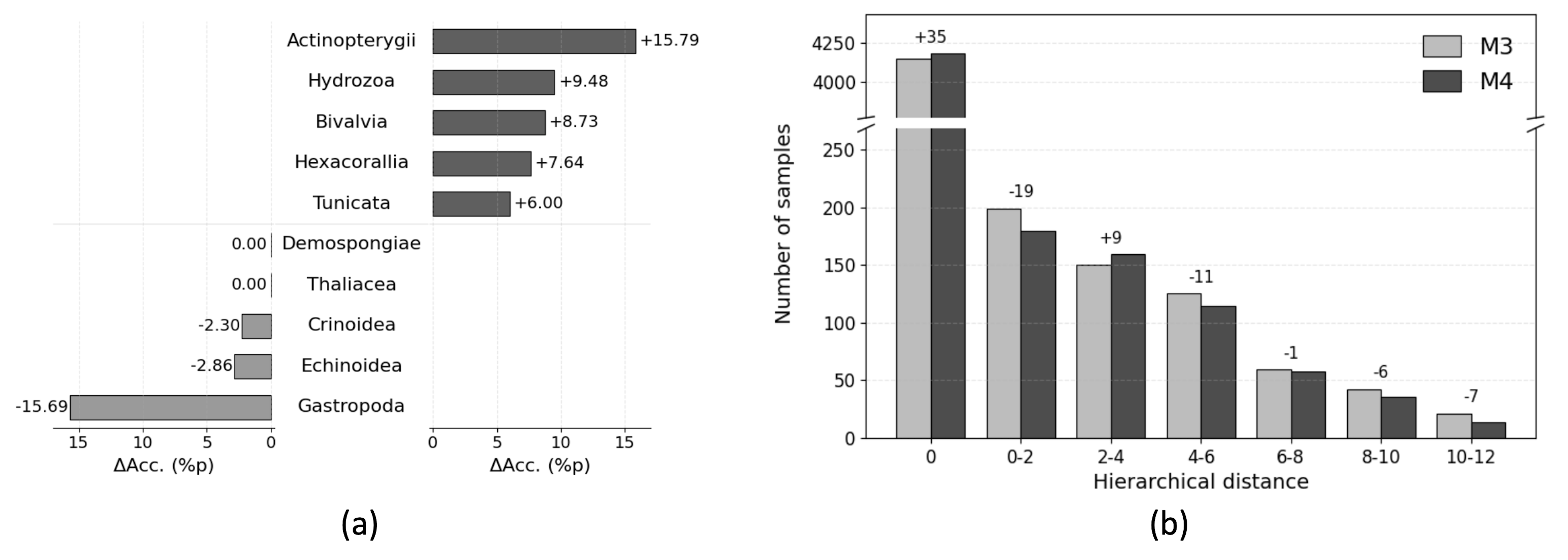}
\caption{Effects of multi-scale contextual views and taxonomy-aware auxiliary supervision on the FathomNet 2025 validation set. (a) Differences in finest-label classification accuracy between M3 and M0, computed separately for samples grouped by taxonomic class. $\Delta$ Acc. is defined as $\Delta \mathrm{Acc.} = \mathrm{Acc.}_{M3} - \mathrm{Acc.}_{M0}$. The five taxonomic classes with the largest gains and the five with the lowest $\Delta$ Acc. values are shown. (b) Hierarchical distance distributions without and with LW-CE as taxonomy-aware auxiliary supervision. M3 denotes the model without auxiliary supervision, whereas M4 denotes the model with LW-CE. Bars indicate the number of samples in each hierarchical distance range.}
\label{fig:context_tax_aux}
\end{figure*}

Figure~\ref{fig:context_tax_aux} provides further analysis of the effects of multi-scale contextual views and taxonomy-aware auxiliary supervision on the FathomNet 2025 validation set. Figure~\ref{fig:context_tax_aux}(a) shows the differences in finest-label classification accuracy between M3 and M0 for samples grouped by taxonomic class. Among the 19 taxonomic classes, accuracy improves in 14 after incorporating multi-scale contextual views. This suggests that broader views can provide useful information about overall body shape, relative scale, and surrounding habitat that may not be fully captured by the target ROI alone. Notably, substantial gains are observed for Hydrozoa, Bivalvia, Hexacorallia, and Tunicata, which frequently include attached, sessile, or colonial organisms. These findings suggest that multi-scale contextual views can support recognition by providing broader observations of the target or additional environmental information, depending on the characteristics of each taxonomic group. In contrast, Crinoidea, Echinoidea, and Gastropoda show decreased accuracy after the contextual views are introduced. For these groups, broader views may introduce heterogeneous or less relevant visual information that distracts the model from discriminative morphological features within the ROI. Overall, multi-scale contextual views are beneficial for most taxonomic classes, although the effectiveness of contextual information varies across groups. Figure~\ref{fig:context_tax_aux}(b) compares the hierarchical distance distributions of M3 and M4. With LW-CE, the number of exact predictions ($\mathrm{HD}=0$) increases by 35 samples, and the overall distribution shifts toward lower HD values. This shift indicates that taxonomy-aware auxiliary supervision produces predictions that are more consistent with the taxonomic hierarchy, in agreement with the lower mean HD reported for M4 in Table~\ref{table:context}.

\begin{figure*}[t]
\centering
\includegraphics[width=1.0\linewidth]{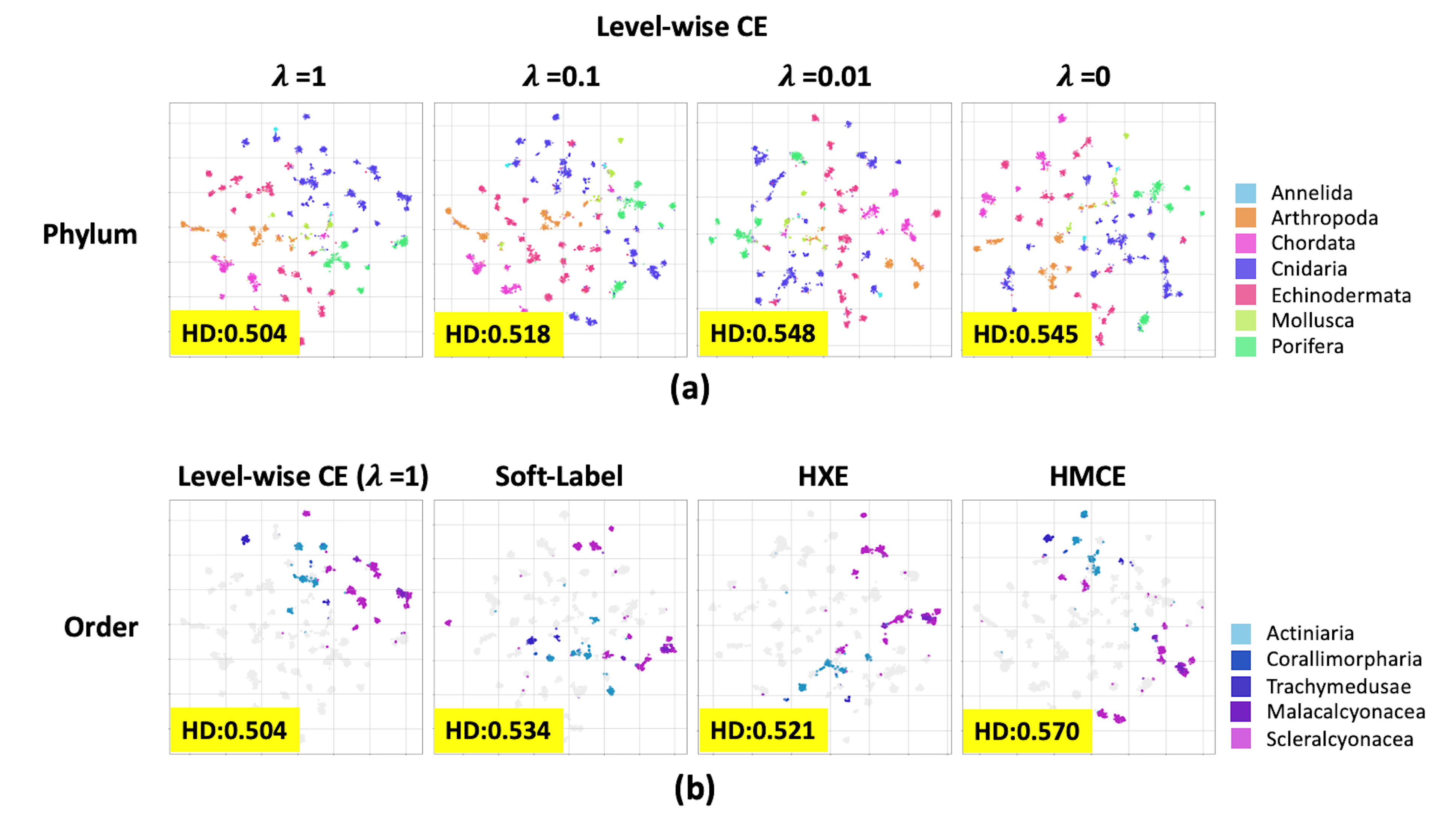}
\caption{t-SNE visualization of the fused embeddings obtained after MCEAM on the FathomNet 2025 validation set. Colors indicate labels at the selected taxonomic rank, and the HD value of the corresponding configuration is shown in each panel. (a) Effect of the LW-CE loss weight $\lambda$, visualized using phylum labels. (b) Comparison of hierarchy-aware auxiliary strategies using order labels within the phylum Cnidaria.}
\label{fig:tsne}
\end{figure*}

To further examine whether taxonomy-aware auxiliary supervision encourages more structured representation learning, we visualize the fused embedding $z$ obtained after MCEAM using t-SNE~\citep{RN40}. In Figure~\ref{fig:tsne}, samples are colored according to labels at the selected taxonomic rank, and each panel reports the HD score of the corresponding experimental configuration. Figure~\ref{fig:tsne}(a) shows the effect of the auxiliary loss weight $\lambda$ using phylum labels. When $\lambda=0$ or $\lambda=0.01$, samples from the same taxonomic group are broadly dispersed in the projected space. As $\lambda$ increases, samples sharing the same taxonomic label form more compact clusters, particularly when $\lambda=1$. This pattern suggests that increasing the contribution of auxiliary taxonomic supervision makes taxonomic grouping more apparent in the projected space and is accompanied by lower HD scores. Figure~\ref{fig:tsne}(b) compares hierarchy-aware auxiliary strategies at the order rank within Cnidaria. SoftLabel and HXE produce more visually distinguishable local groupings than HMCE, whereas the proposed LW-CE produces relatively compact groupings and achieves the lowest HD among the compared strategies. Overall, these visualizations provide qualitative evidence that auxiliary taxonomic supervision influences local grouping patterns in the projected embedding space. The differences in these grouping patterns are accompanied by differences in HD scores, suggesting an association between the taxonomic organization of the learned representations and hierarchical recognition performance.

\begin{figure}[t]
\centering
\includegraphics[width=1.0\linewidth]{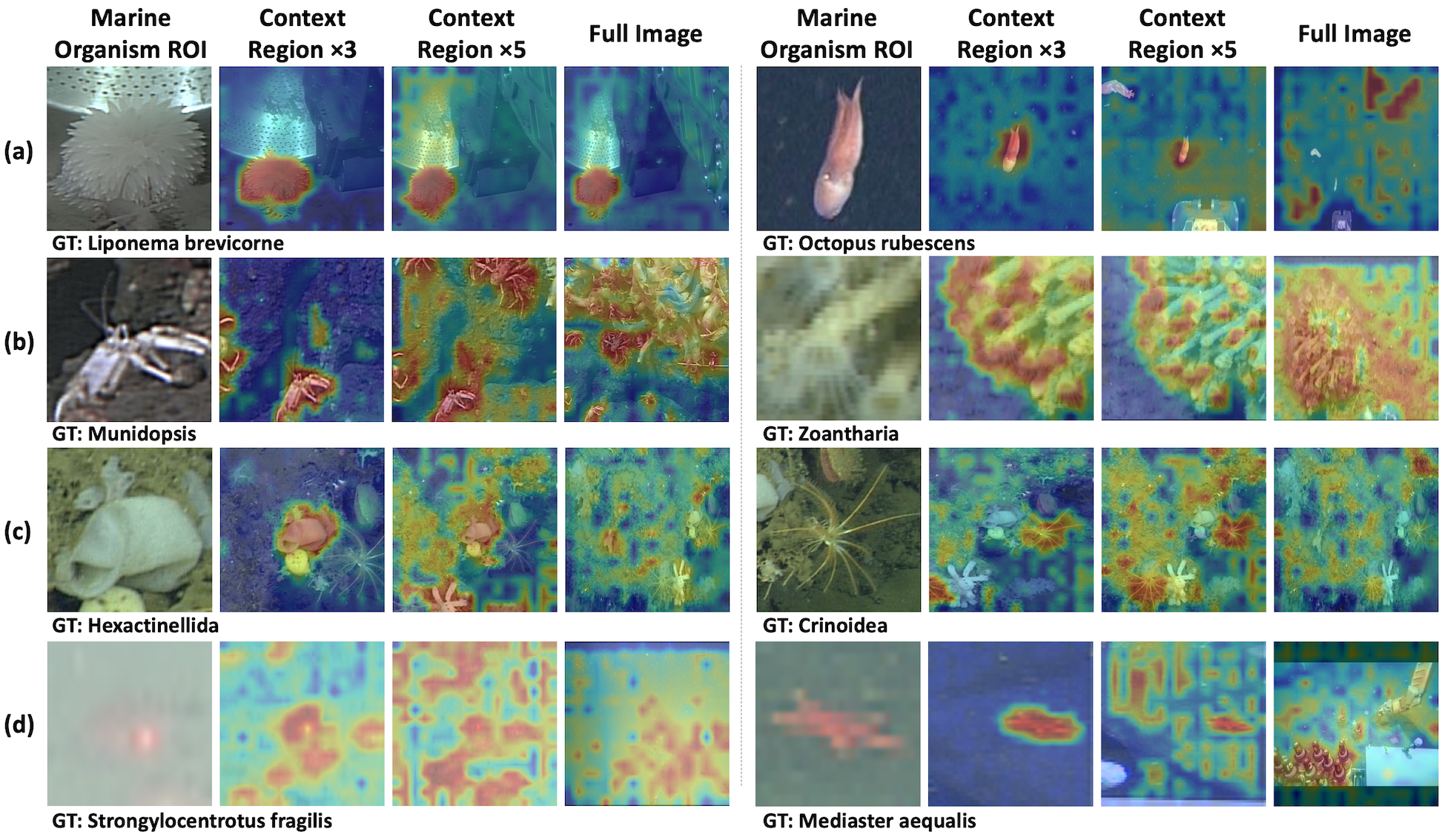}
\caption{Target-conditioned cross-attention maps generated by MCEAM in M4 across different contextual scales on the FathomNet 2025 validation set. The maps indicate spatial regions receiving relatively high attention conditioned on the specified ROI representation and illustrate diverse aggregation patterns across examples.}
\label{fig:attn}
\end{figure}

Figure~\ref{fig:attn} visualizes the target-conditioned cross-attention maps generated by MCEAM in M4 at different contextual scales. The maps indicate the spatial regions in each contextual view that receive relatively high attention conditioned on the corresponding ROI representation. Through this ROI-conditioned attention mechanism, MCEAM adaptively emphasizes target-related cues and surrounding environmental information according to their relevance to the specified ROI, resulting in different spatial attention patterns across examples. In Figure~\ref{fig:attn}(a), relatively concentrated attention is observed around the target organism. For example, the attention maps for Liponema brevicorne highlight portions of the organism, while comparatively lower attention is assigned to some artificial structures. Figure~\ref{fig:attn}(b) and (c) show more spatially distributed patterns. In the examples of Munidopsis and Zoantharia, attention extends to adjacent individuals and nearby colonies. The Hexactinellida and Crinoidea examples further demonstrate that the spatial attention pattern changes with the specified ROI, even when the contextual image remains the same. Figure~\ref{fig:attn}(d) presents less clearly localized patterns. In visually ambiguous or low-resolution examples such as Strongylocentrotus fragilis, attention is distributed over broader image regions. In the Mediaster aequalis example, relatively high attention is also assigned to the recording equipment. This pattern may reflect the repeated co-occurrence of the organism and the equipment in the imagery, allowing the equipment to serve as a salient contextual cue. Overall, these examples show that MCEAM adaptively aggregates target-related and environmental cues across contextual scales according to the specified ROI. The diverse attention patterns demonstrate its ability to capture both localized target information and broader contextual structures.

\subsection{Evaluation Beyond Underwater Imagery}

\begin{table}[t]
\centering
\small
\setlength{\tabcolsep}{3pt}
\renewcommand{\arraystretch}{1.05}
\begin{tabular}{lcccc}
\toprule
\textbf{Method} & \textbf{Acc.} $\uparrow$ & \textbf{Macro-Prec.} $\uparrow$ & \textbf{Macro-F1} $\uparrow$ & \textbf{HD} $\downarrow$ \\
\midrule
VGG19~\citep{RN30}      & 0.667 $\pm$ 0.002 & 0.512 $\pm$ 0.003 & 0.477 $\pm$ 0.005 & 1.236 $\pm$ 0.008 \\
ResNet50~\citep{RN31}   & 0.708 $\pm$ 0.001 & 0.564 $\pm$ 0.002 & 0.529 $\pm$ 0.004 & 1.083 $\pm$ 0.002 \\
ResNeXt50~\citep{RN32}  & 0.668 $\pm$ 0.001 & 0.534 $\pm$ 0.002 & 0.506 $\pm$ 0.002 & 1.235 $\pm$ 0.005 \\
Swin-B~\citep{RN017}    & 0.724 $\pm$ 0.001 & 0.613 $\pm$ 0.003 & 0.579 $\pm$ 0.002 & 1.028 $\pm$ 0.003 \\
MaxViT-T~\citep{RN34}   & 0.687 $\pm$ 0.001 & 0.545 $\pm$ 0.005 & 0.521 $\pm$ 0.004 & 1.162 $\pm$ 0.005 \\
DINOv2-B~\citep{RN33}   & 0.723 $\pm$ 0.000 & 0.610 $\pm$ 0.014 & 0.573 $\pm$ 0.006 & 1.033 $\pm$ 0.002 \\
TransFG~\citep{RN36}    & \underline{0.733} $\pm$ 0.001 & \textbf{0.638} $\pm$ 0.003 & \textbf{0.621} $\pm$ 0.003 & \underline{1.020} $\pm$ 0.005 \\
INTR~\citep{RN35}       & 0.706 $\pm$ 0.005 & 0.553 $\pm$ 0.006 & 0.465 $\pm$ 0.023 & 1.081 $\pm$ 0.023 \\
MSDBN~\citep{RN15}      & 0.671 $\pm$ 0.008 & 0.483 $\pm$ 0.012 & 0.421 $\pm$ 0.021 & 1.199 $\pm$ 0.040 \\
MATANet~(M4)        & \textbf{0.738} $\pm$ 0.002 & \underline{0.630} $\pm$ 0.007 & \underline{0.606} $\pm$ 0.003 & \textbf{0.980} $\pm$ 0.005 \\
\bottomrule
\end{tabular}
\caption{Classification results on the FAIR1M v2.0 remote-sensing dataset. Models are evaluated using accuracy, macro-precision, macro-F1 score, and hierarchical distance (HD). The best and second-best results were determined using the unrounded values and are shown in bold and underlined, respectively.}
\label{table:fair1m}
\end{table}

To further examine the applicability of MATANet beyond marine organism recognition, we evaluate it on FAIR1M v2.0, a fine-grained remote-sensing dataset. As shown in Table~\ref{table:fair1m}, MATANet achieves the highest accuracy and the lowest HD among the compared models. The performance differences between MATANet and TransFG are relatively small and vary across metrics. TransFG achieves the second-best accuracy and the best macro-precision and macro-F1 score, indicating that it remains a strong fine-grained classification baseline in this remote-sensing setting. In contrast, the lower HD achieved by MATANet suggests that its predictions are more consistent with the FAIR1M class hierarchy.

The relatively small performance differences may be related to the characteristics of FAIR1M. Unlike FathomNet 2025, which provides a richer biological taxonomy, FAIR1M has a shallow two-level class hierarchy, limiting the amount of hierarchical information available for auxiliary supervision. In addition, the coarse superclasses in FAIR1M can exhibit substantial differences in both object appearance and surrounding scenes. For example, ships and aircraft are commonly observed in maritime and airport environments, respectively, making contextual information potentially useful for distinguishing between superclasses. However, fine-grained subclasses within the same superclass often share highly similar surrounding scenes while differing mainly in subtle object-level appearance. Consequently, multi-scale contextual views may provide limited additional information for distinguishing among these subclasses. Under these conditions, the benefits of multi-scale context modeling and taxonomy-aware auxiliary supervision may be less pronounced. These results indicate that MATANet remains competitive beyond underwater imagery, while its relative advantage depends on the availability of informative contextual and hierarchical cues.

\subsection{Post-Detection Classification on Matched Object Instances}

\begin{table}[t]
\centering
\small
\setlength{\tabcolsep}{6pt}
\renewcommand{\arraystretch}{1.08}
\begin{tabular}{lcccc}
\toprule
\textbf{Classification Method} 
& \textbf{Acc.} $\uparrow$ 
& \textbf{Macro-Prec.} $\uparrow$ 
& \textbf{Macro-F1} $\uparrow$ 
& \textbf{HD} $\downarrow$ \\
\midrule
YOLO26m native classifier & 0.828 & 0.817 & 0.814 & 0.941 \\
YOLO26m + M0             & 0.861 & 0.852 & 0.853 & 0.673 \\
YOLO26m + M4             & 0.894 & 0.890 & 0.894 & 0.444 \\
YOLO26m + M4-L           & 0.959 & 0.958 & 0.958 & 0.160 \\
\bottomrule
\end{tabular}
\caption{Post-detection classification performance on detector-generated ROIs matched to ground-truth object instances. All methods were evaluated on the same set of matched instances.}
\label{table:det}
\end{table}

The main experiments use ground-truth ROIs to evaluate fine-grained recognition under controlled localization. To examine whether the recognition models remain effective when ROIs are obtained through automatic localization, we additionally evaluate post-detection classification using detector-generated regions. YOLO26m~\citep{RN310} was trained for 100 epochs on the same training split used in the ablation experiments, with an input resolution of $1024 \times 1024$. The detector and recognition models were subsequently evaluated on the disjoint validation split. MATANet was directly applied to the detector-generated regions without additional training or fine-tuning on predicted bounding boxes. A confidence threshold of 0.001 was used to retain a broad set of candidate detections and prioritize localization recall. Predicted boxes were matched to ground-truth instances using class-agnostic greedy one-to-one IoU matching with a threshold of 0.5. Predicted class labels were not considered during matching. For each ground-truth instance, the unmatched predicted box with the highest IoU was selected, and each ground-truth instance and predicted box was used in at most one matched pair. Among 4,743 ground-truth instances, 4,213 were matched to detector predictions, corresponding to a matching recall of 88.8\%, while 530 ground-truth instances remained unmatched. All classification methods were evaluated on the same set of 4,213 matched detector-generated regions, and the classification metrics were computed only on these matched instances.

Table~\ref{table:det} reports classification performance on detector-generated ROIs matched to ground-truth instances. Compared with the native YOLO26m classification output, YOLO26m + M0 improves accuracy from 0.828 to 0.861 and reduces HD from 0.941 to 0.673, demonstrating the benefit of applying a dedicated fine-grained classifier after localization. YOLO26m + M4 further improves accuracy to 0.894 and reduces HD to 0.444, indicating that the multi-context and taxonomy-aware design remains effective when applied to detector-generated ROIs. YOLO26m + M4-L provides an additional improvement, achieving the highest accuracy, macro-precision, and macro-F1 score, together with the lowest HD of 0.160. Overall, these results show that MATANet maintains strong classification performance on detector-generated ROIs, despite potential localization errors. MATANet therefore complements the detector by providing more accurate and taxonomically consistent predictions for successfully localized objects, with further gains obtained through backbone scaling.

\begin{figure*}[t]
\centering
\includegraphics[width=\linewidth]{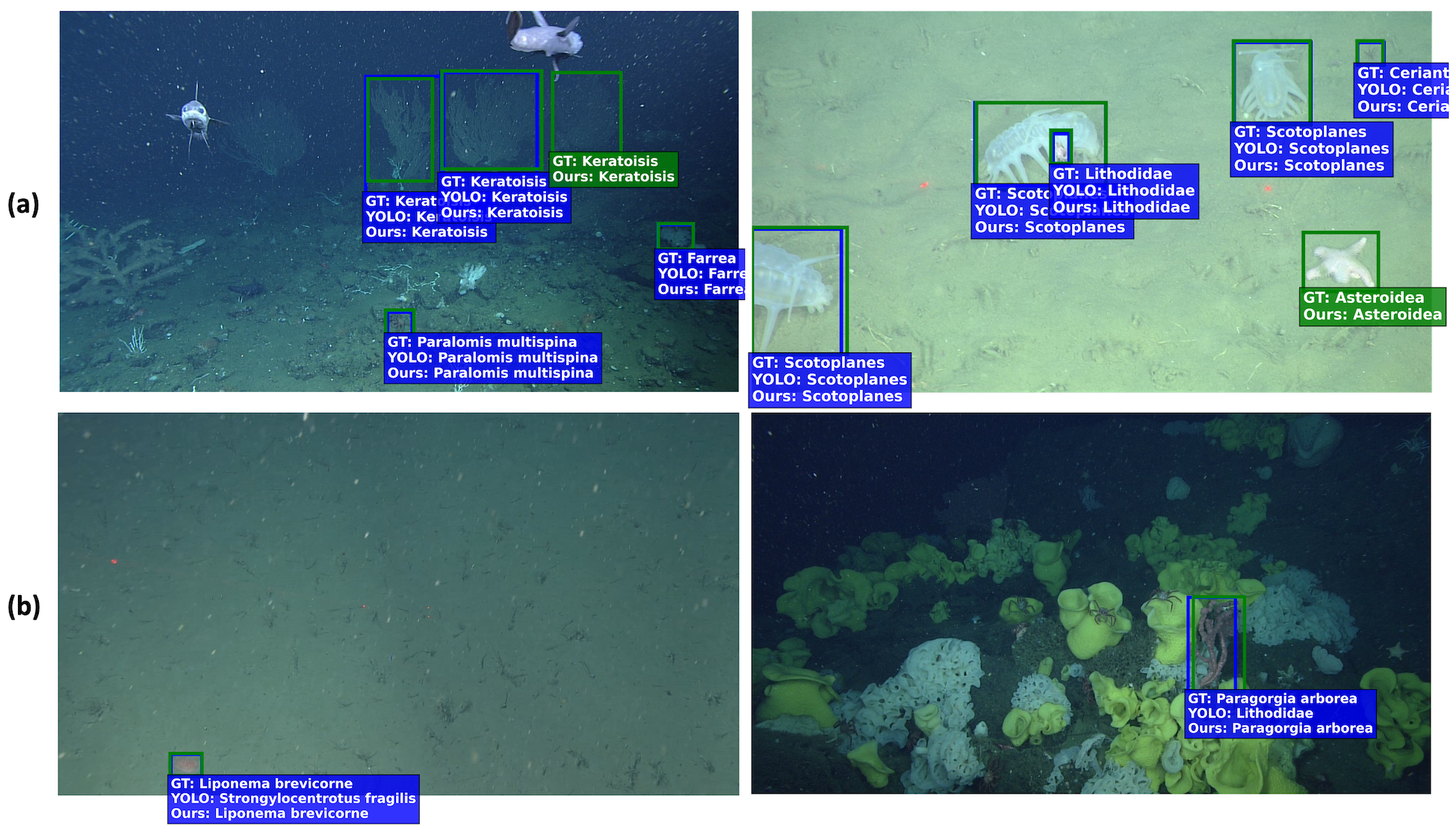}
\caption{Qualitative examples of detection and subsequent fine-grained classification. Green boxes indicate ground-truth objects missed by YOLO26m and are annotated as (GT, MATANet), where the MATANet prediction is obtained using the corresponding ground-truth ROI for qualitative comparison. Blue boxes indicate detector-generated regions matched to ground-truth instances and are annotated as (GT, YOLO26m, MATANet). (a) Examples of organisms missed during automatic localization. (b) Examples in which MATANet corrects or improves the fine-grained classification output of YOLO26m.}
\label{fig:det}
\end{figure*}

Figure~\ref{fig:det} provides qualitative examples illustrating the behavior of MATANet (M4-L) in the post-detection classification setting. Figure~\ref{fig:det}(a) presents organisms that are difficult to localize automatically, including visually inconspicuous or partially occluded instances labeled as Keratoisis and Asteroidea. These examples illustrate that successful localization remains an important prerequisite for fully automatic recognition. MATANet also supports user-assisted recognition: when the detector fails to localize an organism, a user can specify a bounding box around the target, allowing M4-L to provide a fine-grained taxonomic prediction for the selected region. Figure~\ref{fig:det}(b) shows cases in which YOLO26m successfully localizes the target organism but assigns an incorrect fine-grained label. In such cases, M4-L serves as a dedicated post-detection classifier by independently classifying the detector-generated ROI. For example, YOLO26m classifies an instance of Liponema brevicorne as Strongylocentrotus fragilis, whereas M4-L recovers the correct label. Similarly, YOLO26m assigns Lithodidae to an instance of Paragorgia arborea, while M4-L correctly identifies the organism. These examples illustrate a practical detection--recognition workflow in which the detector provides automatic localization and MATANet performs fine-grained taxonomic identification. Together with the quantitative results, these qualitative examples demonstrate that MATANet can complement a general-purpose detector by providing more reliable taxonomic predictions for localized organisms.

\section{Discussion}
\label{sec:discussion}

The experimental results show that fine-grained marine organism recognition benefits from combining the target ROI with multi-scale contextual views. The context-content ablations further indicate that this benefit cannot be attributed solely to repeated observations of the target at different relative scales. Both masking variants underperform M4, indicating that surrounding scene content and repeated observations of the target within the contextual views both contribute to recognition. The larger degradation caused by background removal suggests that surrounding scene information provides particularly important complementary cues in this setting. The aggregation ablation shows that direct concatenation of the independently encoded ROI and contextual embeddings achieves relatively strong performance, indicating that access to multiple contextual views accounts for a substantial portion of the performance gain. Nevertheless, under the same contextual views, backbone, and supervision configuration, MCEAM achieves better results across all reported mean metrics, improving accuracy from 0.869 to 0.881 and reducing HD from 0.568 to 0.501. Beyond these numerical improvements, MCEAM uses the ROI representation to aggregate spatial contextual patches, providing an explicit target-conditioned mechanism whose aggregation behavior can be inspected through cross-attention maps.

The results also suggest that biological taxonomy provides useful supervision for fine-grained recognition. Many marine taxa exhibit subtle inter-class differences, while classification errors may occur at different levels of taxonomic severity. As a result, training only with finest-level class labels does not explicitly account for the hierarchical relationships among taxa or distinguish between taxonomically close and distant errors. By incorporating taxonomy-aware auxiliary supervision while preserving the original fine-grained prediction space, MATANet achieves predictions that are more consistent with the biological hierarchy, as reflected by the reduced HD.

The post-detection experiments further support the applicability of MATANet as a recognition component following automatic object localization. A dedicated ROI classifier improves upon the detector's native classification output, while MATANet provides additional gains through its multi-context and taxonomy-aware design. Specifically, the native YOLO26m classifier achieves an accuracy of 0.828 and an HD of 0.941, whereas YOLO26m + M4-L improves accuracy to 0.959 and reduces HD to 0.160. MATANet also produces more accurate and taxonomically consistent predictions from detector-generated regions, despite being applied without additional training or fine-tuning on predicted bounding boxes. These findings support the use of MATANet as a modular recognition component in practical detection--classification pipelines for marine monitoring. 

\section{Limitations}
\label{sec:limitations}

Although MATANet shows strong performance across the evaluated datasets, several limitations remain. First, the proposed framework assumes that an ROI containing the target object is available and therefore requires an additional detection or localization module in practical monitoring systems. Although the post-detection experiments evaluate MATANet on detector-generated regions, the quantitative comparison is restricted to predictions matched to ground-truth instances and therefore does not represent complete end-to-end detection performance. Second, MCEAM relies on manually defined context scales, including 3$\times$, 5$\times$, and full-image context, which enable a controlled evaluation of different contextual scopes but may not provide the optimal context range across varying object sizes, dataset characteristics, and imaging conditions. Third, additional contextual views are not always beneficial, as some categories show decreased performance and visually salient but irrelevant scene elements can attract attention. These observations motivate adaptive context selection and more selective integration mechanisms. 

\section{Conclusions}
\label{sec:con}

In this study, we proposed MATANet, a multi-context attention and taxonomy-aware network for ROI-guided fine-grained marine organism recognition. MATANet jointly models ROI-based target representations, ROI-centered multi-scale contextual views, and biological taxonomic relationships. MCEAM uses the ROI representation to perform target-conditioned aggregation of spatially resolved contextual patch features, while level-wise cross-entropy incorporates biological taxonomy through auxiliary classifiers at multiple taxonomic levels. Experiments on FathomNet 2025 and FishCLEF2015 show consistent improvements in fine-grained marine organism recognition. The ablations indicate that surrounding scene information provides complementary cues beyond repeated observations of the target at different relative scales, while taxonomy-aware supervision improves the taxonomic consistency of the predictions. Qualitative analyses further illustrate the learned embedding structure and target-conditioned attention distributions over contextual regions. Additional experiments on FAIR1M v2.0 examine the applicability of the proposed design beyond underwater imagery, and the post-detection experiments show that MATANet improves fine-grained classification on matched detector-generated ROIs. The proposed model improves accuracy from 0.828 to 0.959 without additional fine-tuning, demonstrating its engineering applicability to automated marine monitoring.

\bibliographystyle{cas-model2-names}

\bibliography{references}



\end{document}